\documentclass[sn-mathphys,Numbered]{sn-jnl}% Math and Physical Sciences Reference Style
\usepackage{graphicx}%
\usepackage{multirow}%
\usepackage{amsmath,amssymb,amsfonts}%
\usepackage{amsthm}%
\usepackage{mathrsfs}%
\usepackage[title]{appendix}%
\usepackage{textcomp}%
\usepackage{manyfoot}%
\usepackage{booktabs}%
\usepackage{algorithm}%
\usepackage{algorithmicx}%
\usepackage{algpseudocode}%
\usepackage{listings}%
\usepackage{subcaption}%
\usepackage{xcolor}%
\usepackage{soul}%

\usepackage[section]{placeins}
\usepackage{booktabs}
\usepackage{siunitx} % For aligning numbers at the decimal point

\theoremstyle{thmstyleone}%
\theoremstyle{thmstyletwo}%

\theoremstyle{thmstylethree}%

\begin{document}

\title[High-density Electromyography for Effective Gesture-based Control of Physically Assistive Mobile Manipulators]{High-density Electromyography for Effective Gesture-based Control of Physically Assistive Mobile Manipulators}

\author*[1]{\fnm{Jehan} \sur{Yang}}\email{jehany@andrew.cmu.edu}

\author[1,2]{\fnm{Kent} \sur{Shibata}}

\author[1]{\fnm{Douglas} \sur{Weber}}

\author[1]{\fnm{Zackory} \sur{Erickson}}

\affil[1]{\orgdiv{\orgname{Carnegie Mellon University}, \city{Pittsburgh}, \state{Pennsylvania}, \country{United States}}}

\affil[2]{\orgdiv{\orgname{Johns Hopkins University}, \city{Baltimore}, \state{Maryland}, \country{United States}}}

\abstract{High-density electromyography (HDEMG) can detect myoelectric activity as control inputs to a variety of electronically-controlled devices. Furthermore, HDEMG sensors may be built into a variety of clothing, allowing for a non-intrusive myoelectric interface that is integrated into a user's routine. In our work, we introduce an easily-producible HDEMG device that interfaces with the control of a mobile manipulator to perform a range of household and physically assistive tasks. Mobile manipulators can operate throughout the home and are applicable for a spectrum of assistive and daily tasks in the home.  We evaluate the use of real-time myoelectric gesture recognition using our device to enable precise control over the intricate mobility and manipulation functionalities of an 8 degree-of-freedom mobile manipulator. Our evaluation, involving 13 participants engaging in challenging self-care and household activities, demonstrates the potential of our wearable HDEMG system to control a mobile manipulator in the home.}

\maketitle

\section*{Summary} 

This study introduces a wearable HDEMG interface with real-time gesture decoding for neuromotor control of a mobile manipulator.

\section{Introduction}

Electromyography (EMG)~\cite{atzori2014electromyography, pizzolato2017comparison, pancholi2021robust, sultana2023systematic, ozdemir2022dataset} and high-density electromyography (HDEMG)~\cite{geng2016gesture, jiang2021open, parajuli2019real} have been utilized effectively for gesture recognition. EMG uses individual electrodes placed on specific muscles, while HDEMG captures myoelectric activity across an entire limb using an array of electrodes. This allows HDEMG to recognize a wider range of gestures without the need for precise electrode placement, simplifying the donning process~\cite{geng2016gesture, jiang2021open, parajuli2019real}. Due to these characteristics, HDEMG is a promising interface that can be worn and used daily for the teleoperation of assistive mobile manipulator and caregiving robots. It enables the real-time detection and decoding of complex muscle activity patterns, providing volitional motor control information that determines intended hand and wrist gestures and movements~\cite{farina2017man, ting2021sensing, oliveira2022you, allard2016convolutional, fall2016wireless, lee2023stretchable, tian2019large, farina2023toward, gu2023soft, kim2020deep}. One key advantage of myoelectric control interfacing is its potential to serve as an intuitive and high-performance interface, leveraging the user’s learned hand movements and muscle fiber activity~\cite{lobo2014evaluation, corbett2011comparison, godoy2022emg, twardowski2018motor}. Evidence for the advantages of myoelectric interfaces are demonstrated in both \citet{lobo2014evaluation} and \citet{corbett2011comparison}. When evaluating on a 1D target tracking task, it was found that an myoelectric control interface has the same or superior tracking performance compared to an interface controlled by force sensors around the wrist or hand. For the same task,~\citet{lobo2014evaluation} found better tracking performance for an EMG-based control than joystick-based control. Additionally, a high number of possible gestures have been shown to be able to be decoded with high accuracy using HDEMG in previous work~\cite{parajuli2019real}. 

These gestures have the potential to be used for the control of complex robotic systems with many degrees of freedom. As an additional benefit, HDEMG can be integrated into clothing and could serve as a natural and comfortable interface~\cite{alizadeh2022mass, moin2018emg, roland2019insulated, gao2022use, hahne2018simultaneous}. In previous work, it has been shown that individuals with complete hand paralysis can have the residual neuromotor signals from their forearms decoded into different intended hand gestures using machine learning methods and an HDEMG wearable array~\cite{ting2021sensing, oliveira2022you, meyers2024decoding}. 
Robotic systems such as prosthetics~\cite{cimolato2022emg, jiang2012emg}, manipulators~\cite{wang2012development, ison2015high}, and mobile robots~\cite{bisi2018development, wang2022design} have been controlled using EMG, with additional examples found in a survey by \citet{xiong2024intuitive}. 

Our work is the first to control all actuators of a mobile manipulator using EMG. By mapping intended gestures to the movements of an assistive robot that combines both mobility and manipulation capabilities, our work enables individuals to independently control the full robot using motor intent signals from paralyzed portions of their bodies. This comprehensive control is critical for enabling a range of household and self-care tasks, making our system a significant advancement in the field of assistive robotics.

Teleoperation of mobile caregiving robots has the potential to enable those with motor impairments to perform activities of daily living (ADLs) and household tasks on their own~\cite{iskandar2019employing, vogel2020edan}. These activities include assistive tasks such as self-hygiene, eating, and dressing ~\cite{hetz2009activities}. However, current methods for controlling these robots can be cumbersome and difficult to use for people with motor impairments, often requiring the operator to manipulate buttons and joysticks on handheld controllers. For people with quadriplegia, these handheld interfaces are impractical due to the hand dexterity they demand. While some interfaces utilize residual neck and mouth movements, they can disrupt other concurrent activities, such as speaking or gazing~\cite{padmanabha2023hat, mohammadi2023tongue, chen2001application, mougharbel2013comparative}. This highlights a fundamental issue: people with paralysis face difficulties interfacing with wearable devices that require movement. There is therefore a benefit in developing novel interfaces that do not utilize their limited movements, but instead the motor intent using neuromuscular signals from their paralyzed muscles. Recent research indicates that the residual neuromotor activity in the forearms of people with motor-complete quadriplegia from spinal cord injury (SCI) can be harnessed to decode hand motion intent with significant degrees of freedom via electromyography~\cite{oliveira2022you, ting2021sensing}. This insight holds promise for controlling high degree-of-freedom robots, which could enhance user autonomy. In this paper, we introduce a method for teleoperating mobile caregiving robots. The wearable enables the acquisition of surface HDEMG signals, which are used in a machine learning algorithm to translate myoelectric activity into precise robot control actions to complete tasks in a home setting. 

\begin{figure*}[h!]

  \begin{subcaptiongroup}
    \textbf{A}
    
    \includegraphics[width=1.0\textwidth]{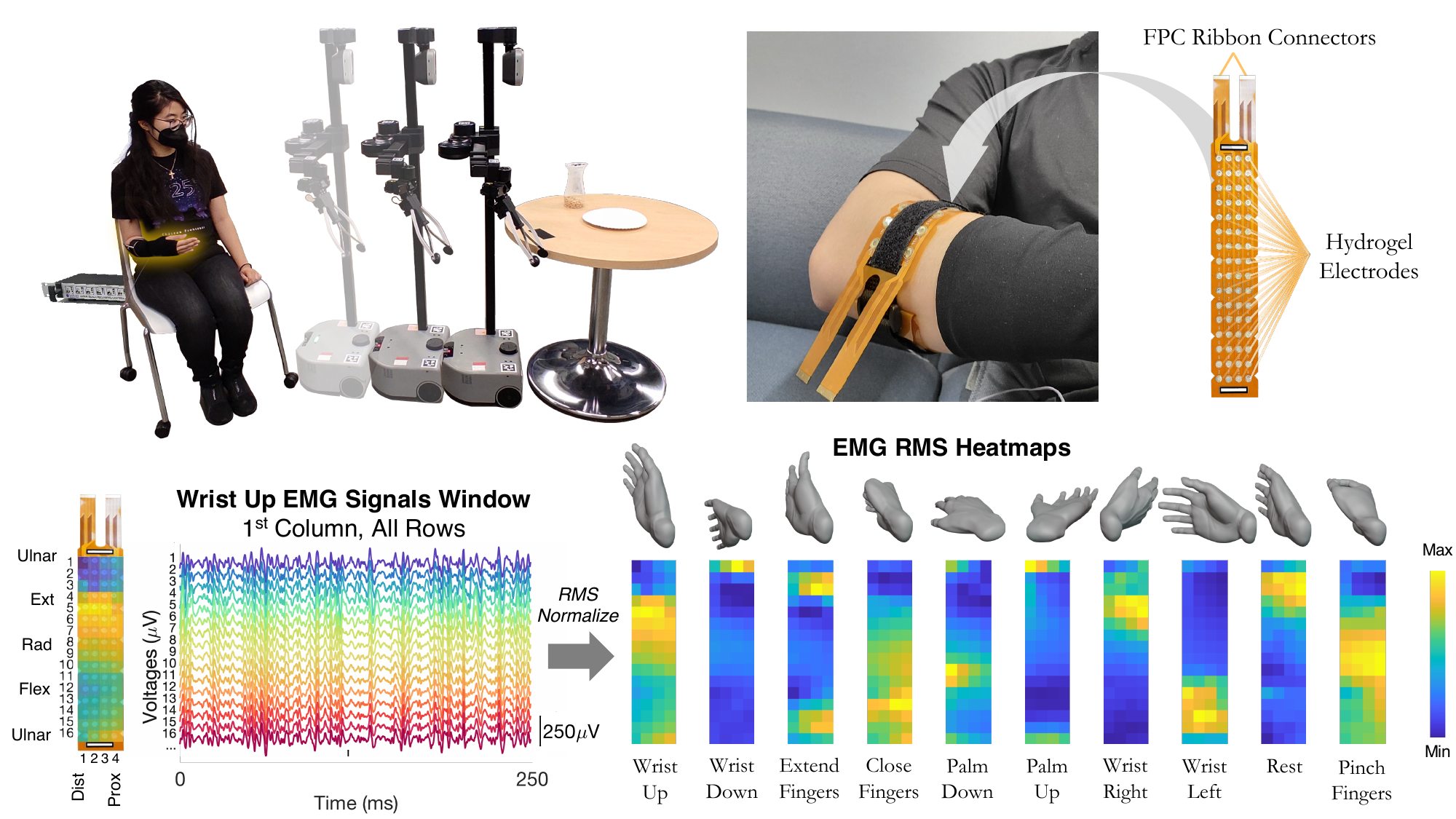}
    \textbf{B}
    
    \begin{center}
    \includegraphics[width=0.7\textwidth]{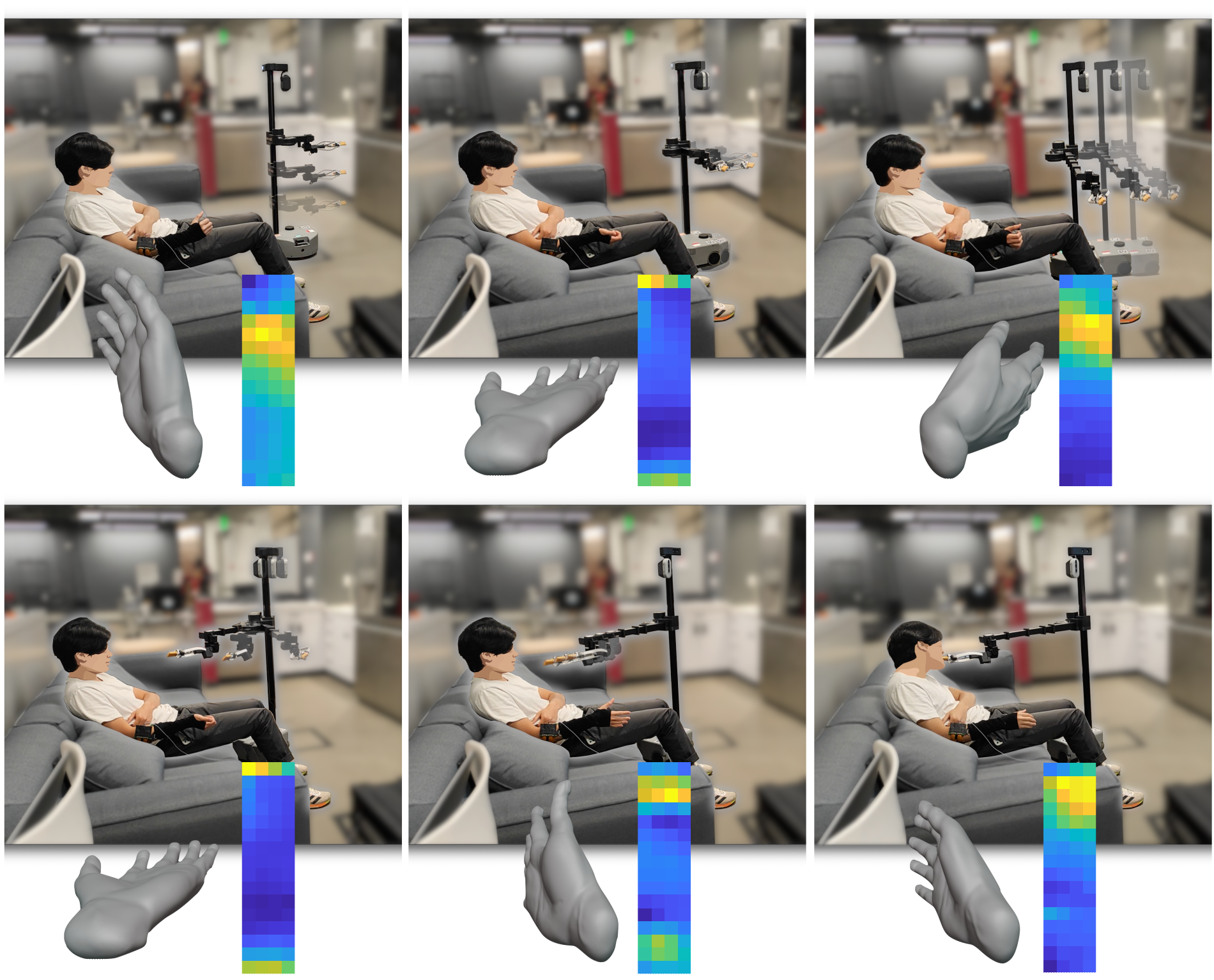}
    \end{center}
  \end{subcaptiongroup}
  \vspace{1em}
  \caption{\textbf{System Overview.} (\textbf{A}) Participant sitting with HDEMG sleeve on chair controlling a Stretch RE2 robot to move towards a table to perform an assistive task. The Intan RHD Recording Controller is behind the user and an SPI cable connects the sleeve to the recording controller. On the right, we show a close-up of participant wearing custom HDEMG sleeve, along with a diagram of an unrolled HDEMG device. Below, we show voltage signals over time seen during gesture performance. To the right of the signals, we show that different gestures have varying patterns of voltage root-mean-square (RMS) heatmaps. (\textbf{B}) These frames show different movements that the participant gestures in order to control the robot to bring a cereal bar close to their face. The corresponding gesture and sample RMS heatmaps from the participant's data are shown for each gesture and movement. }
  \label{overview}
\end{figure*}

In previous EMG decoding work, both EMG devices and decoding algorithms used have varied, resulting in various compromises of both signal quality, and algorithm robustness and complexity~\cite{rani2023surface} EMG devices include ones with individual channels~\cite{atzori2014electromyography, pizzolato2017comparison, pancholi2021robust, sultana2023systematic, ozdemir2022dataset} or arrays of channels~\cite{geng2016gesture, cote2019deep, krilova2019emg, jiang2021open} , and wet electrodes~\cite{jiang2021open, geng2016gesture, ozdemir2022dataset} or dry electrodes~\cite{krilova2019emg, cote2019deep, atzori2014electromyography, pizzolato2017comparison, pancholi2021robust, sultana2023systematic}. Popular methods for decoding algorithms include linear discriminant analysis~\cite{atzori2014electromyography, englehart2001wavelet, chan2007myoelectric, khushaba2014towards}, support vector machines~\cite{atzori2014electromyography, chan2007myoelectric, toledo2019support, khushaba2014towards}, K-nearest neighbor~\cite{atzori2014electromyography, khushaba2014towards}, random forest~\cite{atzori2014electromyography, findik2020random}, convolutional neural networks~\cite{ozdemir2022hand, geng2016gesture, dere2023novel, nahid2020deep, cote2019deep, lin2020normalisation, li2023deep, wang2023pruning, lu2022domain, islam2024surface, chen2020hand, zhai2017self, hu2018novel}, recurrent neural networks~\cite{hu2018novel, wu2018dynamic}, graph neural networks~\cite{lee2023stretchable, xu2023cross}, transfer learning~\cite{cote2019deep, li2023deep, lu2022domain, islam2024surface}, the usage of time-frequency~\cite{ozdemir2022hand, chen2020hand, duan2015semg, zhai2017self, englehart2001wavelet}, blind source separation~\cite{farina2017man, oliveira2022you, ting2021sensing}, or manifold representations of EMG data~\cite{xiong2022learning}. Algorithms also vary depending on whether the output is for discrete movement classification or continuous kinematics estimation, such as enumerated in a review paper by \citet{xiong2024intuitive}. In our work, we create a pipeline involving a neural network that achieves high classification accuracy using spatial representations of the data as shown in Figure~\ref{overview}A, allowing us to visualize and interpret the data.

In our work, we use a wearable HDEMG robot control device designed to capture myoelectric signals and translate them into commands for a home-based mobile manipulator, as shown in Figure \ref{overview}B. This wearable consists of a reusable hydrogel electrode array designed for the forearm, as depicted in Figure \ref{overview}A. This wearable also requires minimal skin preparation, in which we clean the skin and apply a moisturizing lotion. Subsequently, the wearable is donned quickly and secured using a single textile strap. This is in contrast to typical dry HDEMG grids, which usually require significant skin preparation involving hair shaving and skin abrasion, as well as the careful application of an adhesive template and conductive gels to improve impedance for the bare metal electrode contacts. One recent work by \citet{zhang2023stretchable} developed a flexible PCB with semi-permanent hydrogel electrodes for a neck-worn interface using a custom layer-by-layer fabrication approach. In contrast, our work focuses on a forearm-worn interface and employs a simpler fabrication process using commercially available flexible PCBs that can be manufactured by conventional PCB fabrication services, along with standard off-the-shelf hydrogel sheets. During a brief calibration phase, the EMG signals are recorded and processed to train a decoder in minutes after wearing the device. The decoder interprets the signals into specific gestures in real-time; these gestures are subsequently converted into actions for the mobile manipulator robot. To assess the feasibility of our HDEMG control system, we conducted a human study to evaluate its performance in controlling the mobile manipulator. Through this interface, participants were able to execute various self-care and household tasks, such as feeding, meal preparation, and adjusting blankets. These tasks required a blend of mobility, manipulation, and interaction with the human body. Our findings revealed that participants were able to complete quickly using the HDEMG system, along with 12 out of 13 participants agreeing that they were able to effectively use the interface to control the robot, even as they used this novel interface for the first time.  The contributions of our work are threefold: 1) the development of a wearable surface HDEMG device with semi-permanent hydrogel electrodes that can be donned within a few minutes, 2) the first pipeline for myoelectric mobile manipulator control, enabling the decoding of 10 gestures in real-time from myoelectric data and mapping them to an 8-degree-of-freedom mobile manipulator, and 3) the first human study demonstrating the use of myoelectric signals to intuitively control a mobile manipulator to complete household and assistive self-care tasks. These contributions collectively demonstrate the feasibility and effectiveness of full mobile manipulator robot control using wearable EMG sensing.

\section{Results}

\subsection{Design of HDEMG Sleeve}

The wearable surface HDEMG sleeve is designed to not only capture myoelectric activity from muscles around the entire circumference of the forearm, but also to be reusable and applied quickly. The hydrogel electrode HDEMG array, shown on the arm in Figure \ref{overview}A, is made of a flexible, printed, polyimide circuit board (flexPCB) with a total thickness of 0.1mm and is composed of 64 gold-immersion-plated through-holes with sticky conductive hydrogels (Axelgaard AG640), each 0.4mm in thickness attached around each through-hole pad on the flexPCB. These hydrogels are semi-permanent, being able to be kept on between multiple sessions and not requiring the application of a precise grid of liquid gels for every session such as used by existing commercial HDEMG systems~\cite{tmsi_hd_emg_electrodes, tmsi_textile_hd_emg_grids, otbioelettronica}. This allows for the electrodes to be cleaned and rehydrated by wiping with a wet fabric and eliminates the need for separate adhesive and liquid gel layers, simplifying the donning process to wrapping the armband around the user's forearm. The through-holes have an outer diameter of 5.8mm and an inner diameter of 3.2mm. The interelectrode distance is 10.0mm in the direction tangential to the arm and 15.0mm in the direction parallel to the axis of the arm~\cite{gallina2022consensus}. The use of hardened polyimide on the top and bottom layers of the flexPCB supports rectangular openings for a strap to be tightened. Through these openings, we insert a velcro strip with a ladderlock slider buckle to tighten the wearable device around the arm. 

A total of 64 electrodes are used, with the traces going out to two 32 pad flexible printed cable (FPC) connectors supported by hardened polyimide on one side. The FPC connectors are connected to standard zero-insertion force connectors on a breakout board, which connects to a recording controller (Intan RHD Recording Controller) using a standard Intan serial peripheral interface cable. The ground is connected to the back of the hand and the reference is connected on top of the pisiform bone on the ulnar-dorsal side of the wrist, both using standard medical-grade Ag-AgCl gelled electrodes. 

Before the electrodes are attached to the arm, a disposable wipe is used to clean the skin of oil, dirt and residue that could affect the impedance of the skin~\cite{day2002important}. The electrode array is attached to the arm by palpating the ulnar bone and placing the first row of electrodes roughly 1cm in the dorsal direction from where the ulnar bone is felt. The rest of the array is laid over the thickest part of the forearm, approximating the location of the muscle bellies of the largest muscles in the forearm. This donning procedure can be easily accomplished within a few minutes, with the full procedure shown in Supplementary Figure 1. %\ref{supfig:donning}.
We note that a black polyester sleeve is worn over the armband and the ground wire in order to secure the ground electrode wire in place. An advantage of the hydrogels being directly attached to the grid is that it does not require conductive gel or cream to be applied to the skin~\cite{stegeman2012high, lapatki2004thin, afsharipour2016spatial, moin2021wearable} or apply adhesive tapes~\cite{lapatki2004thin, afsharipour2016spatial, stegeman2012high, chandra2020performance} like other HDEMG grids.

\subsection{Characterization}

When employing EMG sensors, it's important to consider the impact of impedance variations over time. These variations can significantly influence the signal-to-noise ratio (SNR) and the patterns of arm signals detected by the sensors. Impedance changes often arise due to factors like the accumulation of sweat and skin debris, as well as physical movement, which can cause electrode sliding and deformation.

To address these variations, we conduct an evaluation of electrode impedance values at the beginning of the study and around an hour afterwards, measuring impedances at 1000 Hz from the hydrogel electrodes to the ground electrode. This second evaluation is concurrent with a recalibration period, involving the collection of new EMG data. We provide additional details in Section \ref{sec:initial_data_collection}. The second impedance data measurement is conducted 67 minutes into the onset of the study, on average, with a standard deviation of 9.9 minutes. Supplementary Figure 2 %\ref{supfig:impedances}
shows the values of the impedances over the 64 electrode array.

 While the overall change in mean impedance values across all electrodes is relatively modest at 1.7\%, we note that a statistically significant difference in impedance values did emerge between the initial data collection ($M = 303~k\Omega, SD=90.5~k\Omega$) and during recalibration ($M = 308~k\Omega, SD=80.1~k\Omega$) as determined by a one-tailed Wilcoxon signed rank test ($W = 132{,}645, n = 832, p < 0.001$). In this statistical test, we define the pairs as the same electrode between the two data collection periods.

In further analysis of changes in impedance, we observe an average percent change in impedance values of 5.2\% for each electrode, with a standard deviation of 24.5\%. While the average percent change in electrode impedance may appear relatively minor, the high standard deviation suggests considerable variability. Although these changes in individual electrode impedances may not significantly affect the overall signal quality, they could decrease the accuracy of gesture recognition for classifiers that rely on consistent patterns of sensed myoelectric activity over time.

 To evaluate this potential decrease in accuracy, we compared the performance of a classifier trained on data from the beginning of the study with its performance on data collected in the middle of the study. We also compared this with a model that is both trained and tested on data from the middle of the study. Details about the data collection process and model training are provided in Section \ref{sec:initial_data_collection}. The resulting average confusion matrixes are presented in Figure \ref{confusionmatrices}, and the confusion matrices of individuals before recalibration are shown in Supplementary Figure 3.%\ref{supfig:confusion_matrices_before_recalibration}.

 However, despite the observed impedance increases in individual electrodes over the course of the study, our results emphasize the robustness of our HDEMG-based control system. Participants are able to effectively complete all teleoperation tasks, even with the shifts in signal distributions from the electrode array that may come from impedance changes. We show the changes in impedances from the initial measurement to the recalibration measurement in Supplementary Figure 2. %\ref{supfig:impedances}. 
 These results underscore the adaptability and reliability of our wearable HDEMG device for long-term use in assistive applications.

The SNR is a critical metric as it quantifies the quality of the signal obtained from the sensors relative to the background noise. We conduct an assessment of the SNR of the sensor array within each participant's setup. To measure SNR, participants are prompted to perform a specific gesture we call ``Fingers Closed.'' During this gesture, participants are instructed to press their fingers into their palms with maximum force, allowing us to capture the maximum voluntary contraction (MVC) signal. Following this, a ``Rest'' gesture is executed to capture the baseline noise level when the muscles are in a relaxed state. These measurements are taken at the beginning of each data collection session.

To calculate the SNR, we divide the RMS of the MVC data by the RMS of the ``Rest'' gesture data, utilizing only the final 2 seconds after participants were cued to hold the gesture. For further detail, the different parts of the gesture cues are visualized in Figure \ref{dataprocessing}. This specific time frame is chosen to eliminate any potential confounding factors associated with the initial moments of the cue, such as varying reaction times among participants. We detail the SNR calculation in this equation:

\begin{align}
    SNR = \frac{\sqrt{\frac{1}{CT}\sum_{c=1}^C\sum_{t=1}^T m_{c,t}^2}}{\sqrt{\frac{1}{CT}\sum_{c=1}^C\sum_{t=1}^T r_{c,t}^2}} = \sqrt{\frac{\sum_{c=1}^C\sum_{t=1}^T m_{c,t}^2}{\sum_{c=1}^C\sum_{t=1}^T r_{c,t}^2}},
\end{align}
where $C$ is the number of channels, $T$ is the number of timesteps, $m$ is the raw data signal during MVC, and $r$ is the raw data signal during the ``Rest'' trial. 

The SNR measurement allows us to assess how effectively the myoelectric signals corresponding to different gestures can be distinguished from background noise. Importantly, our analysis revealed no statistically significant difference in SNR between the initial data collection phase ($M = 5.7,\mspace{1mu}SD=2.1$) and the recalibration data collection phase ($M = 5.9,\mspace{1mu}SD=1.7$), as determined by a Wilcoxon signed rank test ($W = 34.0, n = 13, p = .227$). This lack of significance suggests that the SNR remained consistent throughout the study, with no notable degradation in signal quality. This finding underscores the consistent presence of high signal amplitudes in our system, even given the moderate increase in electrode impedances that we have seen. 

\subsection{Initial Data Collection and Model Training}
\label{sec:initial_data_collection}

At the beginning of the human study, participants undergo a 10-minute data collection session where we capture 5 minutes of gesture-specific data. This session comprises of two series, each containing 50 gestures. Each series consists of five repetitions of each of the ten distinct gesture types, with a one-minute rest period between the series. The ten distinct gesture types are listed in Figure \ref{overview}A. To eliminate potential biases originating from repetitively performing gestures in a fixed sequence, we randomize the order of gestures within these series. We also note that the timing of the ``Rest'' gesture is cued to be performed just as the other gestures, meaning that balancing of the ``Rest'' gesture with other gestures is not needed.

Figure \ref{dataprocessing} shows the gesture cues that are shown to the participant in order to collect data. Each gesture cue is structured as follows: a rest period of $a$ seconds, a transition time of $b$ seconds to move into the cued gesture, a hold duration of $c$ seconds for maintaining the gesture, followed by $d$ seconds to revert to a neutral resting stance. Only the final $c'$ seconds of the held gesture were considered for data collection, negating any transition effects. In this study, $a = 0.5, b = 1.0, c = 3.0, c' = 2.0,$ and $d = 1.0$. The $c'$ seconds of data are separated into non-overlapping samples of 250 ms, which leads to 8 samples per trial. Because we collect 10 trials per gesture and there are 10 gestures, 800 samples are collected during the initial data collection. Some gesture data might be discarded if inaccurately executed, based on either researcher observations or participant feedback. We note that the $a$ second rest period between gesture cues is not primarily intended to prevent fatigue. Instead, it simulates the real-time control of the robot. To prevent fatigue, we incorporate a 1-minute rest period after every 5 minutes of data collection.

Using these data, we train an artificial neural network to classify hand and wrist gestures from HDEMG signals, creating a personalized gesture decoding model. The input to this model comes from a preprocessing pipeline, which processes 250 ms windows of the data from 64 electrodes captured in real-time, explained in further detail in Section \ref{sect:robotmapping}. The neural network utilizes a fully-connected architecture, with 512 nodes per hidden layer and two hidden layers. Dropout is utilized after each hidden layer to decrease overfitting. This model outputs a vector of the probabilities of the ten different gestures in a one-hot encoding format using the softmax activation function. This neural network model has a mean test classification accuracy of 99.6\% over the 13 participants, with a standard deviation of 0.8\%. Our neural networks perform better than a support vector machine (SVM) and random forest (RF) model trained on the same data, which respectively have mean test accuracies of 92.1\% and 98.9\% with standard deviations of 4.9\% and 1.4\%. Additional details on the parameters used for the SVM and RF are in Section \ref{sec:dataprocessing_methods} and the confusion matrices are shown in Supplementary Figure 4.%\ref{supfig:svm_rf_confusion_matrices}. 

However, we do notice a decrease in the model gesture recognition performance after about an hour of continuous use due to an increase in impedance and physical shifting in the electrodes. To evaluate this occurrence, we capture another 2.5 minutes of ground truth data as part of a recalibration dataset. We show RMS heatmaps over the 64 electrodes for a particular participant, as well as a distance matrix and similarity matrix between the initial data and the recalibration data for all participants in Figure \ref{fig:datainitialvsrecalibration}, where lighter values indicate more similar gesture pairs. We use distance matrices and similarity matrices as they are often used to compare the similarity of subsets of data~\cite{khalighinejad2017dynamic, kriegeskorte2008representational, mihalcea2006corpus}. The RMS heatmaps, the distance matrices, and the similarity matrices suggest that the patterns of activation are similar between the initial and recalibration datasets. The distance matrix is generated by calculating the Euclidean distance of the mean RMS heatmap of the samples from one gesture to the mean RMS heatmaps for all 10 gestures from both datasets. Each of these 20 comparisons generates one row. We use the same method to generate a cosine similarity matrix by calculating cosine similarities between gesture data rather than Euclidean distances. Since the cosine similarity matrix is normalized by heatmap magnitudes, it captures the similarity of relative patterns of activation. In addition, the Euclidean distance matrix captures differences in magnitudes as well. In the heatmap visualization in Figure \ref{fig:datainitialvsrecalibration}A, we see the similarity between gestures from the two datasets when looking at the corresponding heatmaps. The distance and similarity matrices in Figure \ref{fig:datainitialvsrecalibration}B show the similarity of gestures within and between datasets. The similarity of the EMG signals for the same gesture between datasets is evident when looking at the lighter colored diagonal lines in the top-right and bottom-left quadrants. Evaluating each participant’s personalized model on this new data captured after continuous use results in an average accuracy of 93.8\% with a standard deviation of 4.7\%. In contrast, gesture recognition performance averaged 99.6\% for participants at the start of the study, representing an average 5.8\% decrease in recognition performance over 1 hour of continued use. Figure \ref{confusionmatrices}B shows the performance change for each participant in the form of classification test accuracies before and after recalibration. 

In order to alleviate the effects of the signal shift, we train a new classifier on only the new data in order to capture the new signal distributions. Given that the classifier is a 2-layer neural network, the full training time for this newly connected data only takes 2 minutes. This results in an increase in performance back to the 99\% range, with an average test classification performance of 98.8\% and a standard deviation of 1.3\%. The effectiveness of this simple recalibration method shows significant improvements to the classification accuracy of HDEMG data using a short recalibration period after an hour of continuous use. We note that there is a 1\% decrease in mean test accuracy across all participants from the initial model to the new model, which may come from the decreased amount of training data. In this case, due to an emphasis on providing a fast recalibration method, half as much data is collected for training. 

To further characterize the classification performance given real-time visual gesture-decoding feedback to the participant, we collect one additional test set for the initial model. This test set is designed to quantitatively evaluate classifier performance in real-time given a gesture matching task in which participants are given gestures to perform. 
Like during the initial data collection, a series of randomly ordered gesture cues are provided with each gesture being cued exactly five times. However, unlike during the first 10 minutes of data collection, real-time visual feedback of the currently classified gesture is provided to the participant using the real-time decoder trained using the initial data.  The participant holds the gesture for at least 3 seconds after the transition period, from which we measure the number of predicted gestures which match the cued gesture. In this time, at least 20 predictions are made before the cue for the gesture holding period ends. We show the results of the mean and median prediction accuracy for each gesture cue from the 1st cue to the 50th cue during the test in Figure \ref{onlineclassification}A. For all gesture cues and all participants, the overall mean classification accuracy for each cue is 95.6\% with a median of 100.0\%.  Interestingly, this is slightly lower than the original 99.6\% accuracy derived from the initial data collection session. We discuss possible reasons for this effect in further detail in Discussion. 
 
To understand if the sequence in which gestures are performed affects real-time accuracy due to factors like fatigue, we again group each gesture cue by position for statistical analysis. Ranging from the 1st to the 50th cue, each position forms distinct groups for examination. Using a Kruskal-Wallis H-test, we find that there are no significant variations in the median real-time accuracies among these sequential groups ($H(49) = 40.1, p = .1157$). 
This particular result diminishes concerns regarding the influence of short-term fatigue on user proficiency and the system's ability to recognize gestures. This accentuates the potential of the system in decoding gestures for robotic control, especially for assistive tasks that span several minutes. 

\subsection{Robot Control Interface}
\label{sec:home-env-evaluation}

Our work introduces the first use of an HDEMG interface for the control of a mobile manipulator robot, enabling participants to execute comprehensive assistive tasks in domestic environments through a neuromotor interface. The overall procedure for configuring gesture decoding and robot control for a participant can be summarized as follows: 1) A helper assists the participant in donning the wearable device on their forearm. 2) We collect myoelectric signal data for the 10 predefined gestures using a gesture cueing system. 3) These collected data are then employed to train a personalized gesture recognition model. 4) The model performs real-time decoding of intended gestures, which enables control of the robot.

After connecting the wearable HDEMG array to the Intan RHD Recording Controller, we are able to interpret the time-varying myoelectric signals collected from the device as gestures are performed by users. We classify the surface EMG signals into 10 classes corresponding to wrist and hand movements. These decoded gestures are subsequently translated into motor commands for the mobile manipulator robot, with individual gestures corresponding to different robot movements. For instance, the ``Fingers Closed'' gesture directly controls the robot's gripper to close and the ``Fingers Open'' gesture directly controls the robot's gripper to open. Intuitive gesture-to-movement mappings such as these significantly aid users in quickly mastering the robot's controls. A comprehensive breakdown of the controls can be found in Section \ref{sect:robotmapping}.

Moreover, the control algorithm's design allows for granular speed control. The control algorithm initiates movements at a slow pace, gradually accelerating to a maximum designed speed as a gesture is maintained, with the robot motor tracking the position and velocity commands using closed-loop PID control. It does so by giving position or velocity commands that increase in their change in magnitude following a power function the longer a gesture is sustained. This design choice ensures minimal movement for gestures performed for short times, but more pronounced movement for sustained ones. As a result, users can effortlessly guide the robot for broad strokes, such as navigating across rooms, or make delicate adjustments like fine-tuning the end effector's position over a food container. More detail on the equations used can be found in Section \ref{sect:robotmapping}. 

To assess our interface's performance, we test its use in various assistive household and self-care tasks. At the heart of this evaluation is the HDEMG robot-control system. The EMG signals are decoded using a data preprocessing pipeline where EMG signals are first filtered of low frequencies that have higher levels of mechanical and powerline noise. Mechanical noise is characterized by movements in the body that can be up to 20Hz, and in cables that can be up to 50Hz~\cite{boyer2023reducing}. Power-line noise occurs at 60Hz in the United States. Very often, power-line noise also introduces harmonics of the fundamental frequency~\cite{boyer2023reducing}. Furthermore, feature extraction is done through the use of the root-mean-square processing step to extract amplitude-based characteristics of the EMG signal. Lastly, further denoising and reduction in dimension is accomplished through PCA. As shown in Figure \ref{overview}A, the RMS processing step allows us to visualize the spatial heatmaps of the EMG signals of various gestures. Figure \ref{dataprocessing} schematically shows the processing steps used in the gesture decoding pipeline, along with further details on the equations used in Section \ref{sect:robotmapping}.

After training a personalized gesture recognition model, participants use this model to control a mobile manipulator to perform a variety of tasks. The purpose of the demonstrated human study is to introduce and assess the feasibility of a wearable HDEMG as a first-of-its-kind standalone teleoperation interface for a high degree-of-freedom mobile manipulator. We assess various factors including task completion times, the usability of the interface (using 7-point Likert scale items), and workload (using a 7-point modified NASA TLX scale). Participants are instructed to sit in one location and minimize movements below the neck and shoulders. These instructions are imposed while the participant teleoperates the assistive mobile manipulator to perform multiple physically assistive tasks.

Participants first perform a practice task consisting of driving to a kitchen counter, grasping a bottle of orange juice, and then delivering that bottle to a side table near the participant. After the practice task, four assistive tasks are performed by the participant. The four tasks are 1) feeding a cereal bar to the participant, 2) meal preparation involving pouring cereal onto a plate, 3) removing a blanket from a participant's legs, and 4) turning a lightbulb until it turns on. The task and associated initial task setups are shown in Figure \ref{tasksetups}. 

The completion times for tasks are recorded and presented in Figure \ref{onlineclassification}B. Individual task times as well as task setups are shown in Figure \ref{tasksetups}, and the number of mode switches for each task and participant is presented in Supplementary Table 1. %\ref{suptab:mode_switches}. 
The task time for the feeding task is the fastest ($M = 3.14 \text{ min}, SD = 1.985 \text{ min}$), while the meal preparation ($M = 3.76 \text{ min}, SD = 1.85 \text{ min}$) and blanket manipulation ($M = 3.97 \text{ min}, SD = 1.88 \text{ min}$) tasks have close mean task times. The task with the longest mean time of completion is the lightbulb task ($M = 5.24 \text{ min}, SD = 2.30 \text{ min}$) due to the additional time needed to align the gripper perpendicular and precisely above the lightbulb. We expect that the fast task time for the feeding task is associated with being able to do the task at a very close distance to the vision of the user, making it easier to align the gripper in the correct place for the task. In contrast, the lightbulb task requires precise placement of the gripper while being a distance away from the participant, making it relatively difficult to align the gripper with the additional difficulty in accurate depth perception. 

We note that only one task took greater than 10 minutes, with a task time of 10 minutes and 27 seconds for the lightbulb task for participant P12. The participant's initial suboptimal approach and an accidental unscrewing of the bulb added to this prolonged duration. The participant also takes an extended amount of time lining up the gripper with the lightbulb. This task performance exemplifies the inherent challenges of this particular task.

After finishing all tasks, participants are asked whether they agree with Likert item statements that correspond to the interface's qualities in controlling the robot in the ``User Experience Survey''. Figure \ref{likertitems}A shows these results. These statements are associated with the user's ability to improve control of the robot, ability to control the robot effectively, intuitiveness of controlling the robot, ease of learning the interface, ability to do tasks in a reasonable time, and ability to convey intentions for robot control. The median response for all but one statement is ``Agree'', with the exception being improvement in robot control, which scores a median of ``Strongly Agree''. The two statements that have any participant disagree are the ones associated with intuitiveness and ease of learning, where participant P07 is the sole participant that disagrees. 

We found that all 13 participants agreed or strongly agreed with the statement S1, ``I was able to improve my ability to control the robot over time.'' This shows participants found they were able to improve their ability to control the robot either through learning better strategies for how to move the robot to accomplish tasks, or improve their ability to work with the robot control interface. This suggests that given sufficient time to practice, for example over multiple days when used as an interface in the home, participants could further improve task performance, especially for a diverse array of self-care and household tasks. 

Figure \ref{likertitems}B displays workload measures using a 7-point NASA TLX Scale. Participants rated the mental demand and effort as high, with median scores of 5 for both. In contrast, the physical demand measure had a lower median score of 3, even if some participants needed to exert additional physical effort for their gestures to be recognized. The strongest median ratings were found in temporal demand, overall performance, and frustration—all rated at 2. While longer-term studies with individuals are necessary to determine if they can reduce mental demand and effort with increased familiarity, as observed in other robot control interfaces, these results underscore the potential of using a neuromotor interface to manage a complex mobile manipulator for first-time users. 

Participants that performed tasks quickly generally agreed with the items pertaining to the ``User Experience Survey'', but also responded to the NASA TLX survey that they had moderate or high workloads. The 5 participants with the fastest task times ranged from average task times of 2.0 minutes to 3.0 minutes. All 5 of these participants agreed or strongly agreed with all the statements in the ``User Experience Survey''. The participants responses to the NASA TLX showed that they had a rating of 3 (low) to 5 (high) out of 7 for mental demand and a 2 (low) to 4 (medium) out of 7 for physical demand in using the interface. Although the participants rated the interface a 4 (medium) to 5 (high) out of 7 for effort, they rated themselves as highly successful, with scores of 1 (very high) to 3 (high) out of 7. 

For participants outside of the top 5 performers, we see a wider range of responses. The range is 2 (low) to 7 (very high) out of 7 for mental workload and a full range of responses 1 (very low) to 7 (very high) out of 7 for physical workload. For effort, these participants also had a larger range, rating themselves from 2 (low) to 7 (very high) out of 7, although similarly they rated themselves as successful with 1 (very high) to 4 (medium) out of 7. The ranges show the trend of less workload for the highest performing participants compared to the other participants. Although perceived workload was high for many participants using the interface, we expect the amount of effort and perceived workload to go down as the users becomes more familiar with the interface as well, which has been seen with increased experience using other interfaces~\cite{abe2019effect, hancock1996effects, leung2010effects}. For example, in Hancock, et al, they demonstrated a drop in the total NASA TLX score of around 40\% after practicing the same 1D compensatory tracking task 10 times~\cite{hancock1996effects}. We also note that only 1 participant disagreed with the Likert items that the interface was easy to learn and that the interface was intuitive, seen in Figure \ref{likertitems}. This suggests that although some participants found the perceived workload to be high when controlling the robot to accomplish tasks using our interface, the participants still generally found the interface to be intuitive and easy-to-learn.

\section{Discussion}

We demonstrate that participants can effectively handle complex assistive tasks using a mobile manipulator and our newly developed wearable myoelectric control interface. The speed at which tasks were completed, combined with feedback from both Likert items and the modified 7-point NASA TLX survey~\cite{ruan2018comparing, padmanabha2023hat, hart2006nasa}, highlights the practicality and user acceptance of our interface. The task times in our study are similar to those from a study involving teleoperation of a Stretch mobile manipulator robot with an IMU-based wearable device~\cite{padmanabha2023hat} in which the average task time for three out of four evaluated assistive tasks is between 5 and 10 minutes each. Although most of the tasks in our work and \citet{padmanabha2023hat} are different, a task that is represented in both works is a task involving removing a blanket off the participant's body. As mentioned in Section \ref{sec:home-env-evaluation}, the average task time for blanket manipulation in our work is 3.97 minutes, while in \citet{padmanabha2023hat} it is 4.35 minutes. However, we note that due to variations in task setups, a direct comparison can not be made, but similar task completion times suggest similar general task times for task completion between these two interfaces. This presents a compelling wearable interface for usage of a mobile manipulator in the home to provide assistance with ADLs. In addition, people with limited hand function due to SCI may be able to use our myoelectric interface to control the robot, because prior work shows that even people with severe SCI can generate gesture-specific patterns of myoelectric activity~\cite{ting2021sensing, oliveira2022you}. By closing the gap between interfaces that utilize intent originating from nerve signals and interfaces that are used for mobile manipulator robot control, this research opens new opportunities for more intuitive and effective assistive technologies.

We recruited our participants from the local population. The intuitive control demonstrated by participants in our study is partly due to the mechanics of the Stretch mobile manipulator. Most of the robot’s joints operate in Cartesian space, meaning that moving the end effector along a single Cartesian axis requires only a single actuator. This prismatic nature of Stretch enables intuitive joint-level control.

In contrast, 6-DoF and 7-DoF table-mounted manipulators, as demonstrated in prior work~\cite{hahne2018simultaneous, ison2015high}, typically require moving all 6 or 7 actuators to move the end effector along a single axis. Joint-based control of the Stretch mobile manipulator has been found to be feasible and intuitive even for participants without prior robot control experience or those with disabilities such as paralysis or spastic hand movements. Previous studies using mouse and IMU-based control interfaces~\cite{padmanabha2023hat, padmanabha2024independence} support this finding. For example, in~\citet{padmanabha2023hat}, one of the two participants with disabilities, who had no prior robot control experience, demonstrated task times comparable to those without disabilities when using joint-based control to operate the Stretch robot. 

Our system demonstrates high accuracy decoding of up to 10 distinct gestures from an array of 64 electrodes embedded in the liner of a cuff that wraps around the forearm. However we note that there is an observed difference in accuracy between the initial data collection session and when the participant is given real-time visual feedback of the decoded gesture. This may be accounted for by a number of factors. As mentioned in Section \ref{sec:initial_data_collection}, there is a drop in accuracy of 4.0\% from 99.6\% to 95.6\%. During real-time feedback, if a participant discerns a gesture misclassification, they may instinctively modify their gesture to correct their myoelectric activity. For example, identifying an error during the transition might lead participants to adjust the effort with which they sustain the gesture. Additionally, sustained misclassifications of a particular gesture could encourage participants to further experiment with muscle activation strategies, trying to adjust so the decoder can classify the correct gesture. The time pressure of the testing environment, combined with the real-time feedback, could cause some participants to perform slightly worse. Generally, given the dissimilar contexts in which the training data and this test set were collected, a slight reduction in accuracy is anticipated. Even a slight dip in accuracy, such as a single misclassification in every 25 gestures, can lead to the $\sim{}$4.1\% accuracy decrease that we see. 

A major challenge in this work is ensuring that the gesture recognition model trained on data captured during data collection will generalize to how the user performs gestures when controlling the robot. This problem is partially illustrated in a quote from P09, who said, ``having some data collection system where you’re also seeing the robot move would be nice in order to map visually what the robot would be doing with your hand gestures.'' P09 had task times that were the second slowest, with an average task time of 5.9 minutes. Similarly, the 4 participants with the slowest task times all mentioned issues of misclassifications during robot control, causing increased task times during robot control even given their high test classification accuracies. Average task times for the 4 slowest participants ranged from 4.9 minutes to 6.1 minutes. In future work, evaluation of how participants may improve their ability to perform gestures more consistently while controlling a robot can decrease the time in which assistive tasks take.

All participants mentioned that most of the gesture mappings to the robot movements were intuitive, where the mapping is shown in Figure \ref{gestures}. However, a mapping that 7 out of 13 participants mentioned they did not find intuitive during the study is the mapping from the ``Palm Down'' and ``Palm Up'' gestures to the respective direction for turning the robot. Even so, participants who had challenges knowing the direction in which the robot will turn still were able to finish tasks quickly, especially since the recovery for turning the wrong way can be reversed quickly by turning the hand in the opposite direction. We expect that with increasing practice with the robot and interface, this mapping can be learned as almost half the participants did not mention facing problems with the rotations being unintuitive, even while using the interface for the first time. Another option in the future is giving users the opportunity to adjust the robot-gesture mapping. This will be particularly helpful as a customization capability for people with quadriplegia using the interface in the long-term, as the gestures that can be decoded consistently may differ depending on the user~\cite{oliveira2022you}.

For any EMG system that decodes a set of gestures using a prediction model, prediction errors may occur over time as signals change due to sweating or drying of electrodes. Other potential causes of signal changes are due to the buckling or shifts in the wearable on the skin over time as gestures are performed. Methods to overcome gesture recognition failures should be considered for more efficient teleoperation. In evaluating our classifier, we find that there are specific misclassifications that occur with relatively high frequency before recalibration. This can be seen in the gesture recognition confusion matrices before recalibration in Figure \ref{confusionmatrices}. On average, The ``Pinch Fingers'' gesture is misclassified most often of all gestures with an average of a 20\% misclassification rate before recalibration. The second and third most misclassified are the ``Palm Up'' and ``Fingers Open'' gestures, occurring at a rate of 12\% and 8\% respectively. We found that there are distinct instructional cues that often help participants perform gestures with lower rates of misclassification. For example, oftentimes participants find the ``Palm Up'' gesture difficult to perform and do not adequately activate muscular signals during this gesture. Emphasizing that this gesture should be performed with additional physical effort helps reduce the rate of misclassification of this gesture as the ``Rest'' gesture. The recalibration also helps significantly, reducing the average error rates for ``Pinch Fingers'' to 4\%, ``Palm Up'' to 5\%, and ``Fingers Open'' to 1\%. Finally, a systematic solution to overcome gesture recognition failures is to use gestures that can be more easily recognizable from other gestures in terms of their muscle activation patterns. Other future steps could entail developing a more robust classification system through the use of the data collected during initial data collection as the training set, and the data collected during recalibration or additional data collection sessions as the test set. A classifier with high test accuracy can help account for the shift in signals over time. 

In our work, we built personalized models that are trained from the user during the start of the study, spending 10 minutes collecting gesture data. In future work, there is an opportunity for training a unified gesture recognition model that generalizes across people using a previously collected dataset. This has been shown in some works which have collected a dataset from 25-30 participants~\cite{ozdemir2022hand, samadani2014hand}. This could eliminate the need to spend time collecting data after donning the device. Another opportunity is to collect a small initial dataset to fine-tune a generalized model, allowing for highly accurate personalized models on substantially less data, for example 1 minute of personalized data~\cite{lin2020normalisation, chen2020hand}.

Direct full teleoperation allows a user to perform a wide array of challenging mobile manipulation tasks in home environments. However, the use of shared autonomy for specific targeted tasks could decrease task times as well as decreased perceived workload, as has been well-documented in prior work~\cite{gopinath2016human}. For assistive robotic interfaces, it has been found that too much assistance decreases people's acceptability of using the robot~\cite{gopinath2016human}. Therefore, when considering shared control with a neural interface, the amount of automated assistance employed in the control loop could be specified according to the preference of the user~\cite{gopinath2016human, makin2023neurocognitive}.

Flexible PCB wearables have the advantage of being rapidly producible and having the flexibility to wrap around curved parts of the body. However, the rigid sheet device cannot stretch around the arm and will not fit all forearms equally well. This causes some electrodes to overlap for smaller arms, and the circumference to not be fully covered for larger arms. This may affect the performance of the decoder for small or large arms. Future designs that conform more closely to arms of varying sizes can improve electrode contact with the skin. For example, a textile wearable can offer even compression of electrodes to the skin~\cite{acar2019wearable}. Thin serpentine connections between electrodes on a flexible PCB can also allow for significant deformations between electrodes~\cite{lee2023stretchable}. Due to muscle atrophy being significant for people after SCI, creating wearable devices that conform to smaller arm sizes will also be important for improved electrode contact with the skin~\cite{castro1999influence}. 

Off-the-shelf conductive hydrogel electrodes allow for rapid manufacturing of an HDEMG wearable. The electrodes we use are mixed with conductive fillers that increase conductivity, however still have relatively high skin-to-electrode impedances due to their small conduct area. Due to drying of the electrodes, we also see moderate overall increases in impedance. In our study, we address electrodes with high impedance by ignoring signals from channels of electrodes that have high skin-to-electrode impedance and therefore low signal quality, with more detail in Section \ref{sec:dataprocessing_methods}. Future opportunities could include the design and use of custom sticky, durable, and low-impedance hydrogel electrodes~\cite{carvalho2020nondrying}, which have lower impedance and may capture more consistent signals over several hours.

Prior research has shown that intended gestures can be decoded readily from people with quadriplegia due to complete SCI \cite{ting2021sensing}, including decoding the fine motor movements of the angles of five individual digits of the hand~\cite{oliveira2022you}. These prior findings in combination with the compelling results from our study suggest a promising opportunity for individuals with SCI. However, future work remains in a thorough in-home evaluation with people with cervical SCI. Additional studies could assess the application of this interface technology applied to mobile manipulator control for participants with quadriplegia and other motor impairments. 

Our personalized decoder models are straightforward to implement and deploy, using a high pass filter, RMS, normalization, and PCA for preprocessing and achieving a 99.61\% test accuracy for the initial model using data collected only during the study. However, we see a slight decrease in classification accuracies before recalibration around an hour after the initial data collection. Further optimization of the gesture classification pipeline may help address this issue, such as through empirically robust representations of EMG data including time-frequency representations~\cite{ozdemir2022hand}, the use of pre-trained convolutional or graph neural networks~\cite{ozdemir2022hand, lee2023stretchable, allard2016convolutional}, augmentations of the training data that better capture variations between gesture performances~\cite{tsinganos2020data,  maksymenko2023myoelectric}, and practical outlier detection methods that can detect significant shifts in EMG data~\cite{zhang2015real}.

Overall, this paper demonstrates the compelling potential of HDEMG control for teleoperating mobile caregiving robots with an intuitive and natural gesture-based interface. A unique HDEMG wearable combined with machine learning techniques to accurately classify gestures in real-time is a key aspect of this work, allowing participants to control the robot to accomplish several challenging tasks, such as feeding, meal preparation, and blanket manipulation. Our results showing the completion of assistive tasks suggest that this approach can offer an integrated wearable interface for accessible teleoperation of mobile manipulators in the home. 

\section{Methods}

\subsection{Signal Processing Hardware}

The HDEMG sleeve is connected by FPC to a breakout board, which connects to an Intan 128 channel headstage, of which we use 64 channels. The headstage is connected to an Intan RHD Recording Controller via a cable that sends SPI signals. The Intan RHD Recording Controller is utilized to record signals with a sampling frequency of 4000 Hz. The Intan RHD Recording Controller also uses a ground and a reference signal, which are both connected to the back of the hand. The Intan RHD Recording Controller is connected via USB to a MacBook Pro, which runs the Intan RHX software to interface with the Intan RHD Recording controller. A socket-based API is used to read EMG signals in real-time to Python, and then is used as input into a preprocessing pipeline and classifier that is implemented in Keras, as detailed in Section \ref{sect:robotmapping}. The computer connected to the Intan RHD Recording Controller communicates to the robot via a wireless LAN connection using a mobile router (TP-Link AC750). Control commands are given using UDP and are given to the robot at a frequency of 6 Hz.

 \subsection{Data Processing Pipeline and Classification Model}
 \label{sec:dataprocessing_methods}

 The EMG data is provided at a sampling frequency of 4000Hz for all channels, where the data are in the form of voltage signals over time. These data are provided by the Intan RHD Recording Controller and sent to the computer for further signal processing and classification. We note that classification-based control is used rather than proportional control in order to evaluate the use of a high-accuracy but discrete classification-based decoding algorithm when controlling a mobile manipulator using HDEMG signals. One significant advantage of discrete classification-based decoding is its potential applicability to individuals with paralysis. It is generally easier to obtain ground truth data on whether a subject with paralysis has attempted a gesture compared to having the subject attempt a gesture to a specific degree. Channels in which the impedances are measured to be greater than 500k$\Omega$ are ignored since very high electrode-to-skin impedances would have significantly lower signal to noise ratios. The resulting number of channels after the channel rejection method is designated as $M$, which is at most 64. 
 
 In real-time gesture decoding, the data are windowed to the last $N$ samples of each channel. In our case, $N = 1000$ and corresponds to the last 250ms of data at 4000Hz. The data processing pipeline first involves the use of a 120Hz 4th order Butterworth high pass filter~\cite{de2010filtering}, a filter defined to have maximally flat frequency responses in the passband. This is utilized to filter out significant noise from movement, and additional noise from nearby elecromagnetic waves from nearby power electronics--powered at 60 Hz in the United States. In this study, we used a higher cutoff frequency than some other studies involving surface EMG~\cite{de2010filtering}. Evidence suggests that high-pass frequencies above 30 Hz, like the one we used, can effectively filter out low-frequency signals that increase with fatigue ~\cite{potvin2004less}. Our empirical results alto indicate high classification performance with this higher cutoff frequency. Additionally, the 120 Hz cutoff frequency helps eliminate noise from the second harmonic of 60 Hz power line noise, which may still be a significant source of noise. 
 
 Our findings show that data preprocessed with a typical 20 Hz high-pass filter achieved an average classification test accuracy of 99.2\%. In contrast, data filtered with a 120 Hz high-pass filter yielded a slightly higher average classification test accuracy of 99.6\%. Although we conducted our human study using a 120 Hz high-pass filter, we believe a 20 Hz high-pass filter would have performed comparably well, given the similarly high test classification accuracy. The filter magnitude equation is as follows: 
\begin{align}
    G(\omega) = \frac{\omega^n}{\sqrt{1 + \omega^{2n}}},
\end{align}
where $\omega$ is the angular frequency and $n$ is the order of the filter. The phase shift for a Butterworth filter is nonlinear with the change in angular frequency~\cite{zumbahlen2009phase}, but is expected with the use of any real-time filter. 
 
Secondly, the RMS of each electrode over the $N$ sample window is taken. RMS reduces the effect of high-frequency noise that may have high magnitudes by rectifying and smoothing out the signal, and is a simple and efficient feature used for surface EMG classification~\cite{spiewak2018comprehensive}:
\begin{align}
x_{j}^{rms} = \sqrt{\frac{1}{N} \sum_{i=1}^N x_{ij}},
\end{align}
where $x_{ij}$ is the filtered data at timestep $i$ for electrode $j$ and $x_j^{rms}$ is the result of the RMS operation for the $j^{th}$ electrode for the last $N$ time steps. 
During training for both the initial personalized model and recalibration model, the data are split into different samples using non-overlapping $N$-sized sliding windows, with a step size of $N$. A 64-16-20 training-validation-test split occurs over these samples, and a z-score transform is taken over the training data in order to first standardize the data over each channel: 
\begin{align}
z_j = \frac{x^{rms}_j - \mu_j}{\sigma_j},
\end{align}
where $z_j$ is the transformed data of channel $j$, $x^{rms}_j$ is the input RMS data sample of channel $j$, and $\mu_j$ and $\sigma_j$ are respectively the mean and standard deviation of RMS data of $N$-sized windows for channel $j$ of the training data. The training data is used to train the model, the validation data is used to provide a separate set of data to evaluate the data at each epoch to select the best model during training, and the test data is used to show the classifier performance on data that is not classified until training is finished. The best model during training is chosen as the model trained at the epoch with the smallest validation loss value. The results referring to accuracy in the results section of this paper refer to test data classifier performance.

Principal component analysis (PCA) over the training data is then used in order to reduce the dimensionality of the data from $\mathbb{R}^{M}$ to $\mathbb{R}^{K}$. We use $K=30$. PCA is performed by first performing a single value decomposition (SVD), with the data matrix $A$, which contains the number of channels $M$ as rows and the number of data points $N$ as columns. The decomposition produces a factorization of the form:
\begin{align}
A = U \Sigma V^T,
\end{align} where the columns of $U$ and the columns of $V$ form two sets of orthonormal bases. To use PCA to reduce the dimensions of the system, the first $K$ columns of $U$ are extracted as the principal components, and a dot product is utilized with each of the columns:
\begin{align}
y_j = \mathbf{z}\cdot \mathbf{u_j},
\end{align}
where $y_j$ is the $j^{th}$ element of the output, $z$ is the $M$-dimension data input, $u_j$ is the $j^{th}$ principal component, and $j$ varies from $1$ to $K$. The new dimension of the data is $K$, which is also the dimension of the input into the feedforward neural network. 

The feedforward neural network with two 512-node hidden layers and dropout layers of 0.2 after each hidden layer is used as the classifier. The output is a softmax activation function over 10 dimensions corresponding to the 10 gestures. These gestures are enumerated in Figure \ref{overview} with the data pipeline architecture visualized in Figure \ref{dataprocessing}. The softmax is defined as: 
\begin{align}
    p_i = \frac{e^{s_i}}{\sum_j^C e^{s_j}},
\end{align}
where $s$ is the output scores of the last layer, which are also called logits. The softmax function is utilized to normalize the outputs of the neural network layer into vector elements in the range $(0,1)$ and which sum up to $1$ into a vector of probabilities. 
Training occurs for 200 epochs with a batch size of 32 using the Adam optimizer, where the model at the epoch with the lowest cross-entropy validation loss is kept for evaluating the test accuracy and for real-time control, and with the categorical cross-entropy loss defined as:
\begin{align}
    CE = -\log(\frac{e^{s_p}}{\sum_j^C e^{s_j}}),
\end{align}
where $s_p$ are the logits for the correct class, and $s_j$ is the neural network score for class $j$. 

 As mentioned in Section \ref{sec:initial_data_collection}, we train support vector machine models and random forest models as well to compare performance with neural network models. The SVM and RF models are implemented in the scikit-learn library in Python. The support vector machine models we test have a regulariziation parameter of 1.0, a radial basis function kernel, and a gamma of $1/K$, where $K=30$ is the number of input features. The random forest models have a max depth of 8, 100 estimators, a minimum of 2 samples required to split an internal node, a minimum of 1 sample to be at a leaf node, 5 for the max number of features to consider for the best split, and with bootstrapping turned on.

\subsection{Mapping to Robot Control}\label{sect:robotmapping}

At some time step $t$, a trained gesture recognition model outputs a 10-dimensional vector of gesture probabilities according to a 250~ms window of HDEMG signals. This prediction is then used to update a confidence array using exponential filtering to maintain gesture prediction consistency over time, where $\alpha$ times the softmax output is added to the current confidence array:
\begin{align}
    p'(t+1) &= (1-\alpha) p'(t) + \alpha p(t),\\
    p_i'(0) &= 1/10, \nonumber
\end{align}
where $p(t)$ is the softmax output from the neural network at time $t$. We select $\alpha$ to be $0.5$. 

If any gesture probability in the confidence vector is greater than some threshold $r$, then the classifier appends this gesture to an $m$-dimensional buffer array, which holds the last $m$ predictions of the gesture. Otherwise, if no element in the confidence vector is greater than $r$, then the ``Rest'' gesture is appended to the $m$-element buffer array to begin commanding the robot to stay stationary. Using simple majority voting, a gesture that is predicted more than $m/2$ of the last $m$ predictions stored in the buffer array is output as the actual gesture prediction. The use of simple majority voting is used to increase the consistency in gesture predictions over time for robot control. In our system, we set $r = 0.5$, and $m = 3$. Given the combination of 1) predictions occurring at a rate of around $6$Hz, 2) the prediction threshold of $0.5$, 3) exponential filtering used to maintain gesture prediction consistency, and 4) simple majority voting where $2$ out of the last $3$ predictions need to be a certain gesture, it takes at least $0.5$ seconds or $3$ predictions by the decoder network for the robot controller to give motion commands in response to a new gesture. We found that this latency did not provide a significant issue for mobile manipulator teleoperation due to the moderate speeds in which the robot was moving. We also provide a look at the decoder output and the decoder results after the exponential filtering, thresholding, and simple majority voting in Supplementary Figure 6.%\ref{supfig:decoder-output}.

There are two different modes for robot control: 1) fine wrist movements and gripper control, and 2) arm movement and base driving. The robot used is the Hello Robot Stretch RE2, which is a lightweight, relatively inexpensive mobile manipulator robot~\cite{kemp2022design}. The robot has been demonstrated to be successful in physically assistive tasks and home environments in other work~\cite{padmanabha2023hat, ranganeni2023evaluating, puthuveetil2022bodies}.

In the wrist and gripper mode, the gestures that map to the movements of the robot are one-to-one with the wrist of the robot. For the "Rest" gesture, the robot does not move. The gestures ``Fingers Closed'' and ``Fingers Open'' close and open the gripper of the robot, respectively. In addition, ``Wrist Left'' and ``Wrist Right'' causes yawing of the end effector, ``Wrist Up'' and ``Wrist Down'' causes pitching of the end effector, and ``Palm Down'' and ``Palm Up'' causes rolling of the end effector. These commands are used to command servo motors, which are given position commands using a PID control loop.
Holding the ``Pinch Fingers'' position for three seconds is used to switch modes. After the mode switch, a cooldown occurs in order to prevent the occurrence of unintentional consecutive switching of modes.

In the arm and driving mode of the robot, the gestures ``Wrist Up'' and ``Wrist Down'' moves the robot arm up and down, respectively. ``Fingers Closed'' and ``Fingers Open'' causes the robot arm to retract and extend, respectively. ``Wrist Right'' and ``Wrist Left'' causes the robot to drive forwards and backwards in a linear direction, respectively. ``Palm Down'' and ``Palm Up'' causes the differential drive robot to turn counter clockwise and clockwise, respectively, corresponding to a rotation in place. These commands are all used to drive stepper motors, which are given velocity commands that are followed using a PID control loop.   
A visual representation of the robot mapping described in this subsection is shown in Figure \ref{gestures}. 

The movement commands for various joints on the robot ramps up to a maximum speed with the amount of time a gesture is predicted consecutively:
\begin{equation}
r(t) = 
    \begin{cases} 
      ax^{1.5} & x\leq k \\
      ak^{1.5} & x > k,
   \end{cases}
\end{equation}
where $r$ is a velocity or change-in-position command for the robot joint, $a$ is a scalar modifier depending on the joint, $x$ is the number of times that the currently predicted gesture has been predicted consecutively, and $k$ is a maximum number of consecutive predictions. In the case of this study, $k = 50$, corresponding to around 8 seconds. We note that whenever the robot is commanded to move a joint that is different from the joint commanded a time step before, the joint from the time step before is commanded to stop. The robot wrist and gripper were given change-in-position commands due to the limitations of the robot API when we ran our experiments. The robot arm and wheels were given velocity commands, which we found made for smoother actuation. A exponent of $1.5$ is utilized to allow the robot joint to speed up quickly, along with allowing the user to have relatively fine control of a robot joint if $a$ is set to a low number and the gesture command is given over a short time span. The exponent also de-emphasizes the effect of a misclassified gesture moving the robot too significantly if the command is given for a short period due to the smaller slope at lower values of $x$. These set points, $r$, are given to the robot controller for the servo or stepper motors, which have inbuilt functions for performing closed-loop PID commands to give current commands to the motors. PID control commands are of the form: 
\begin{align}
    u(t) = K_pe + K_i\int_{i=0}^t e + K_d \dot{e}, 
\end{align}
and $K_p, K_i,$ and $K_d$ are constants that are typically tuned for the motors to minimize the error $e$ between the desired state and the actual state (either position or velocity), while $u$ is the torque command to the motor.
The robot gripper opening and closing is the only joint that we did not have a ramp-up speed mapping. For each time the gesture corresponding to gripper opening and closing is predicted, a constant change in position command is given, as pilot studies identified this to be a more effective control approach for the closing of the gripper. 

\subsection{Robot Task Experiment Setup}

Here, we describe in more detail the setup for the robot task experiments. After a personalized gesture decoding model is trained and the participant practices with the real-time decoder and visual feedback from a screen, the participant watches a pre-recorded video tutorial in which they learn mappings from the gestures to the robot movements. After this video tutorial, we have the participant begin practicing control of the robot through the performance of a practice task. The task involves moving a bottle of orange juice from a distant kitchen counter to a nearby side table, as seen in Supplementary Figure 5. 
%\ref{supfig:practicetask}. 
The participant is allowed to ask any questions during this practice task to familiarize themselves with robot control. 

During all the tasks after the practice task, the participant receives minimal feedback from the researcher in order to simulate an environment in which the user is controlling the robot mainly by themselves. The minimal feedback that we allow is given when informing the participant that the robot is at a joint limit and cannot move the joint further since the user may be unfamiliar with the physical limitations of the robot.

Here, we describe the setups of the task in further detail. Each task is set up so that the robot is in a stowed position at the beginning near the participant. The stowed position involves the robot arm being fully contracted, along with the arm being brought to a low position, and can be seen in Figure \ref{tasksetups}. Participants are sitting stationary on a couch and use hand and wrist gestures to control the robot using the wearable HDEMG. The goal conditions for each task is made clear to the participant before the participant begins any particular task. For the feeding task, a cereal bar is placed in the robot gripper with the gripper initially closed. The task is finished when the cereal bar is brought to the person's face and they are able to take a bite of the cereal bar while remaining seated. For the blanket removal task, the task starts with the blanket on the legs of the participant. The task is accomplished when the blanket is removed from the participant's legs completely. For the meal preparation task, a carafe of cereal is placed next to a plate. The task is finished when any amount of cereal is poured onto the plate.  For the lightbulb turning task, the lightbulb begins already partially socketed into a powered lamp and the task is accomplished when the participant fully screws the lightbulb into the socket and the bulb illuminates.  The blanket task, meal preparation task, and feeding task are evaluated in a counterbalanced random order, while the lightbulb task is always evaluated last due to the increased difficulty of the task. 

In order to account for task failures while still allowing the participant to finish a task, the participant is allowed to ask to restart a task, which pauses the task time while the researcher reset the task to the original state. In our study, resets occurred for 4 of the 52 total tasks done. 

\subsection{Human Participants Research}
Before inclusion, participants gave their written informed consent, including photography and video, and agreed that this material can be used in journals and other public media. Participants were above the age of 18 and recruited through the use of recruitment posters around the Carnegie Mellon University campus. Eligible participants are fluent in written and spoken English, and do not have histories of motor, cognitive, or visual impairments that are not correctable by conventional means. Participants also did not have a history in the past year of neurological or musculoskeletal disorders, or experienced pain or deficiencies in their arm or hand. The study protocol was approved by the Carnegie Mellon University Institutional Review Board, protocol 2021.00000121.

\section{Data Availability}
The data and the code used and/or analyzed during the current study are available from the corresponding author on request.

\section*{Acknowledgments}
We thank J. Song and M. R. Carneiro for their advice on EMG skin electrodes. We also thank L. Nguyen for helping with attaching the hydrogels to the flexPCBs. \textbf{Funding:} This work was supported by the National Science Foundation Graduate Research Fellowship Program under Grant No. DGE2140739 and by the National Science Foundation under Grant No. 2341352. \textbf{Author contributions:} J.Y., K.S., D.W., and Z.E. designed the study. J.Y., K.S. fabricated devices for the study. J.Y., K.S. collected the data. J.Y. analyzed the data. J.Y., Z.E., and D.W. interpreted the data and performed the literature search. J.Y. created the figures. Z.E., and D.W. gave feedback for the figures. J.Y. wrote the manuscript. J.Y., K.S., Z.E., and D.W. edited the manuscript. \textbf{Competing interests:} Authors declare that they have no competing interests. %\textbf{Data and materials availability:} All data needed to evaluate the conclusions in the paper are present in the paper or the Supplementary Materials.

\backmatter

\bibliography{main-text}% common bib file

\clearpage

\section*{Figure Legends}

\begin{figure*}[!]

    \begin{subcaptiongroup}
    
    \textbf{A}
    
    \includegraphics[width=1.0\textwidth]{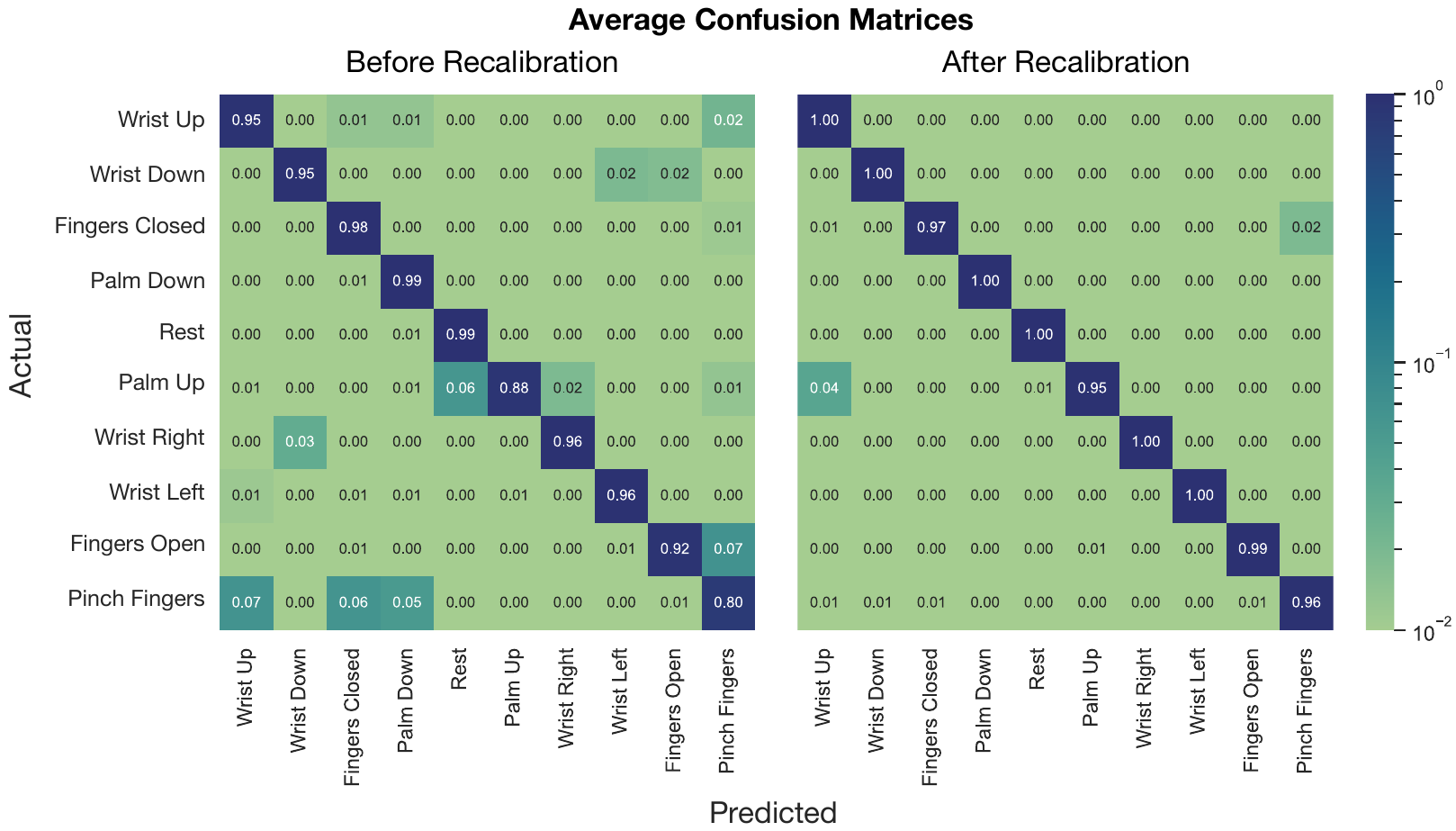}
    
    \textbf{B}
    
\vspace{-1em}\hspace{0.1\columnwidth}\includegraphics[width=0.8\columnwidth]{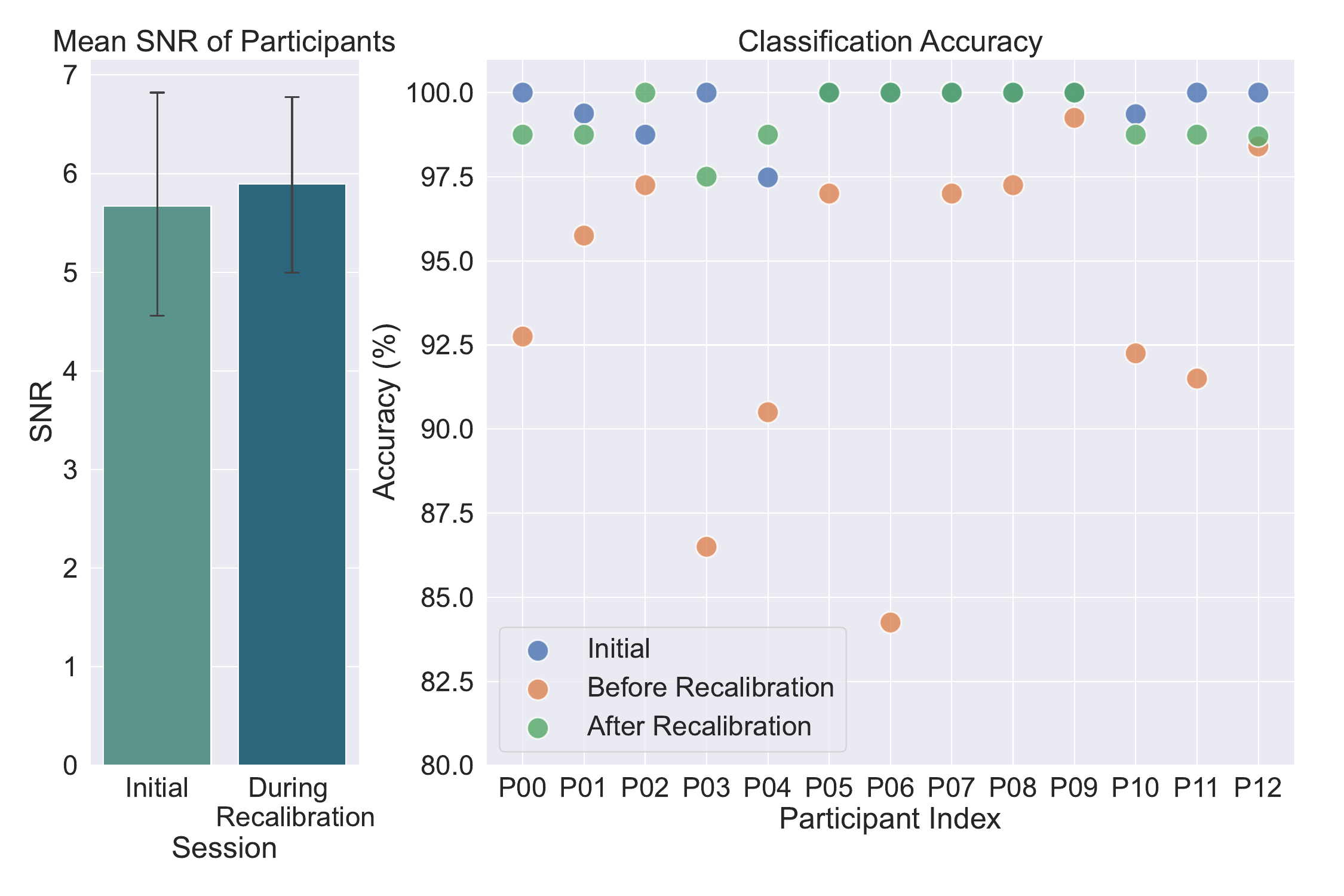} 
    \end{subcaptiongroup}
    \caption{\textbf{Confusion Matrices, SNR, and Test Accuracies. }\textbf{(A)} Test accuracy confusion matrices before recalibration and after recalibration. The confusion matrix ``Before Recalibration'' shows some of the most commonly misclassified gestures after the shift in signals distributions have occurred. The ``After Recalibration'' matrix shows that accuracies are increased after the recalibration phase. \textbf{(B)} Differences between the initial session and the recalibration session. The mean SNR between the initial session and the recalibration session was not found to be statistically significant. The error bars show 95\% confidence intervals for the mean. This shows the test accuracies for the initial dataset, before recalibration, and after recalibration for each participant.}
    \label{confusionmatrices}
\end{figure*}

\begin{figure*}[!]
    
%     \begin{subcaptiongroup*}

%     \textbf{A}
    
%     \includegraphics[width=\textwidth]{images/DataCollection.pdf}
%     \par\medskip

%     \textbf{B}

% \includegraphics[width=1.03\textwidth]{images/DataPreprocessing.pdf}
%     \end{subcaptiongroup*}
    \caption{\textbf{Data Collection and Decoder Architecture. }\textbf{(A)} Data collection GUI, with the first three frames showing a priming phase in which a translucent animation demonstrates the gesture to be performed next. The second set of frames shows an animation of the actual gesture performance where the participant is cued to perform simultaneously. The end of the second row shows the gesture being held, which is held for 3 seconds. The third set of frames shows the gesture returning to neutral, along with a gesture prompt change in text at the top to the next gesture to be performed. \textbf{(B)} Data processing and machine learning pipeline. Raw EMG data is filtered with a high pass filter at 120 Hz. A root mean square is then applied to extract magnitude information for each electrode. A z-score normalization is applied, followed by PCA for dimension reduction. The transformed input is then fed into a feedforward neural network for training and predicting gestures out of 10 classes. }
    \label{dataprocessing}
    
\end{figure*}

\begin{figure}[!]

\textbf{A}

\includegraphics[width=0.9\columnwidth]{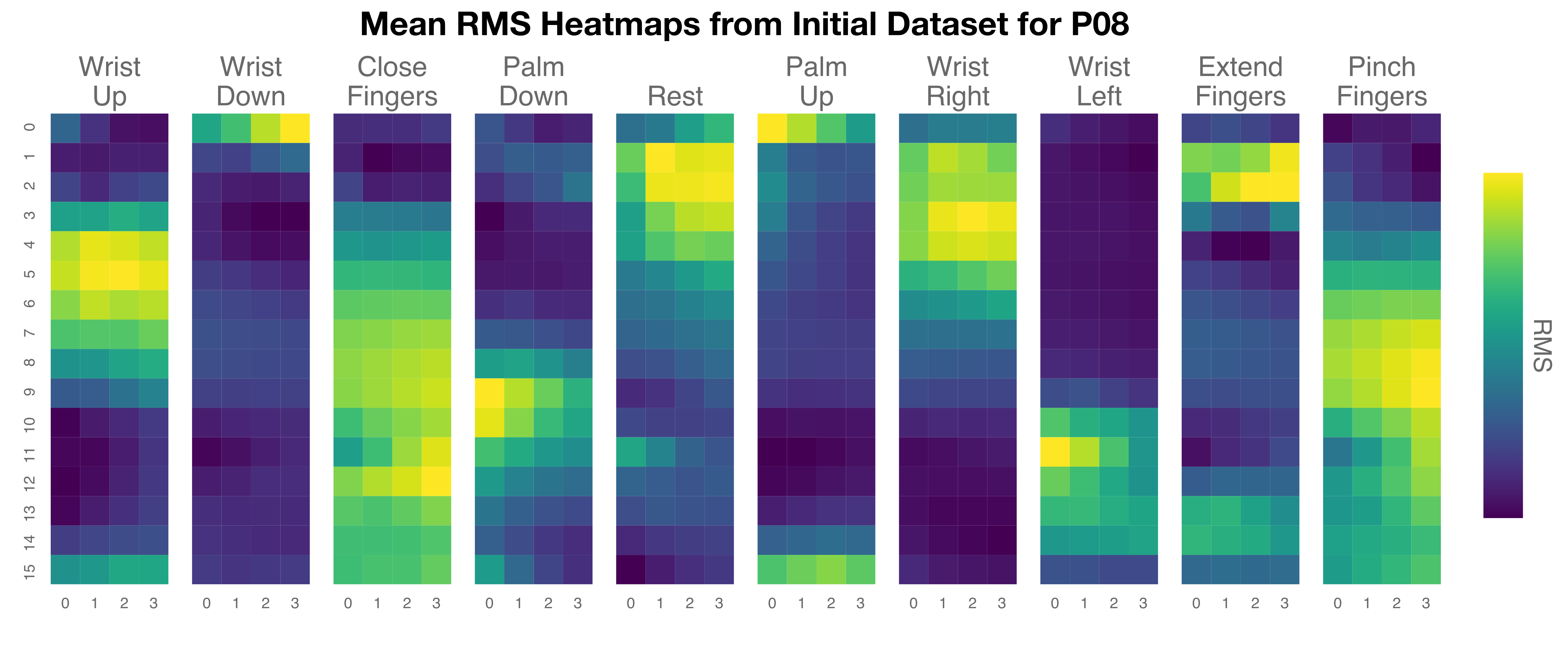}

\includegraphics[width=0.9\columnwidth]{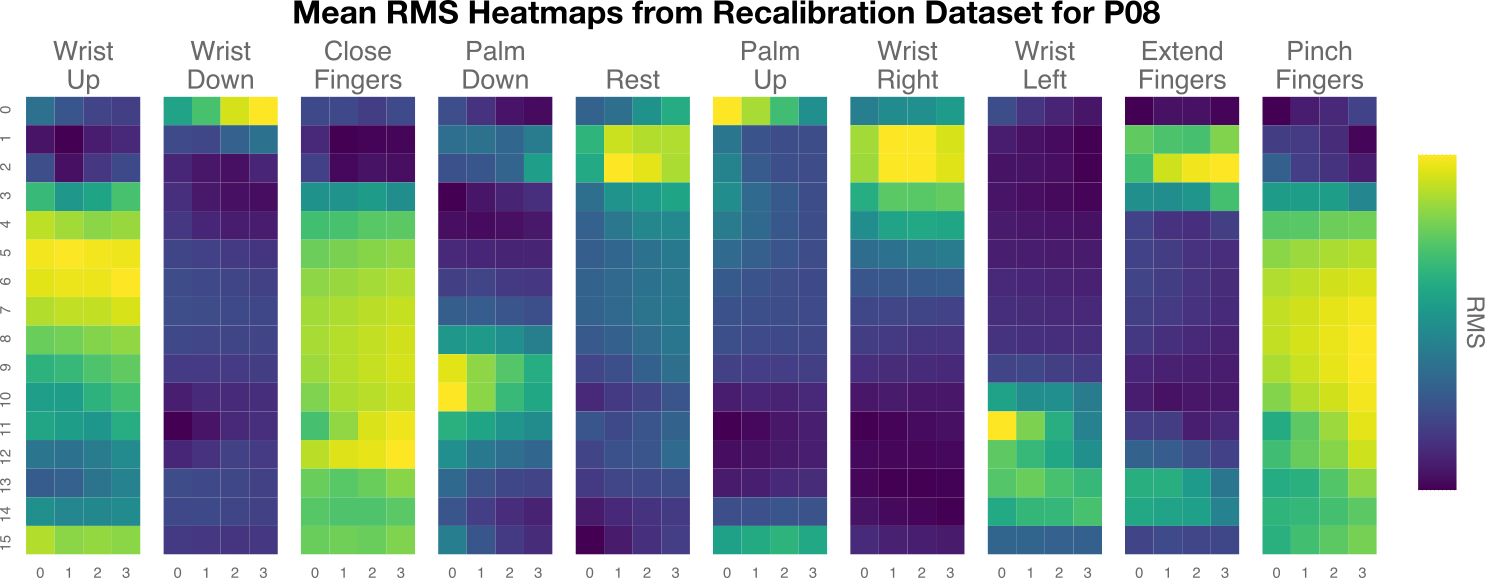}

\textbf{B}

\includegraphics[width=0.475\columnwidth]{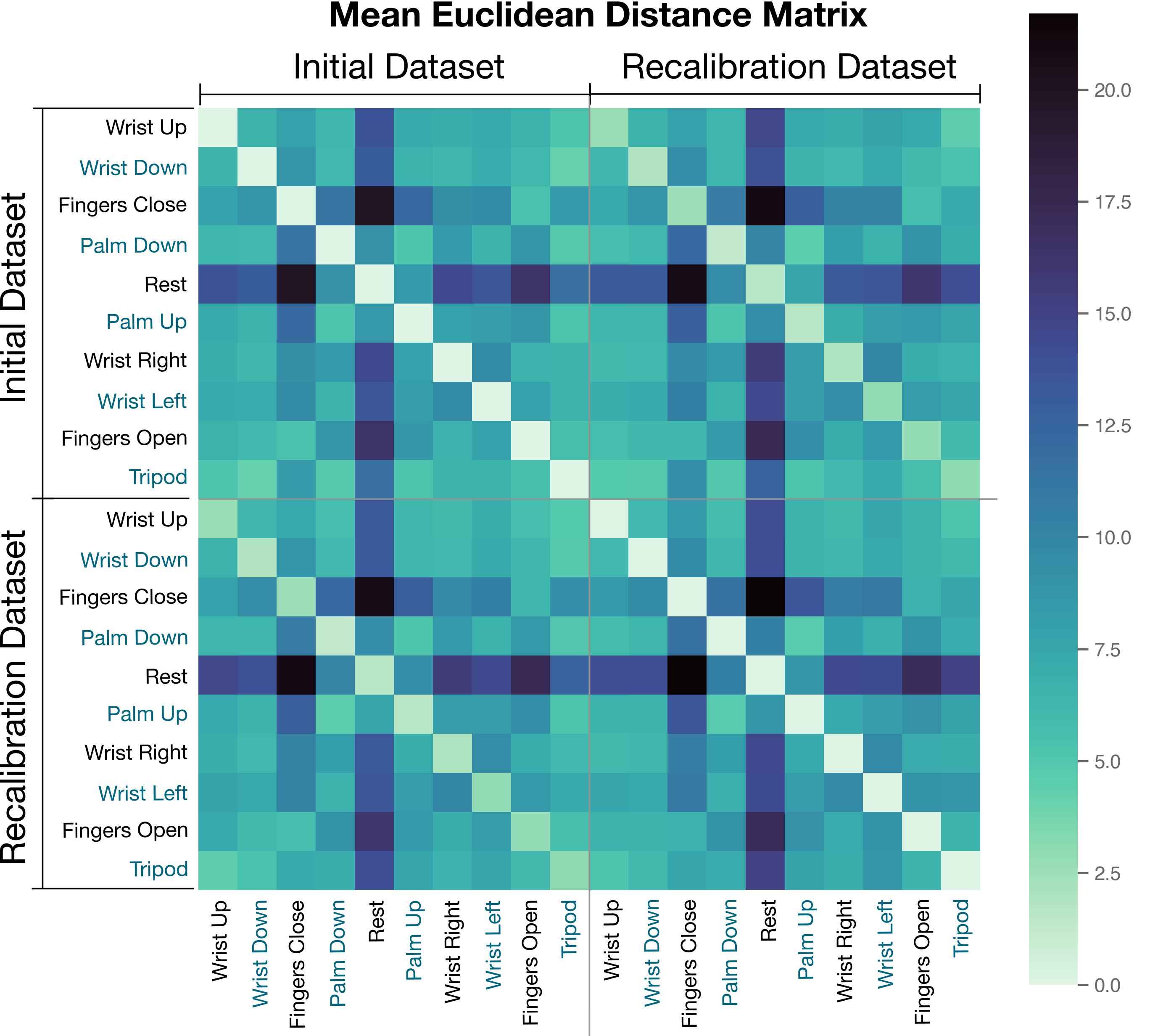}
\includegraphics[width=0.475\columnwidth]{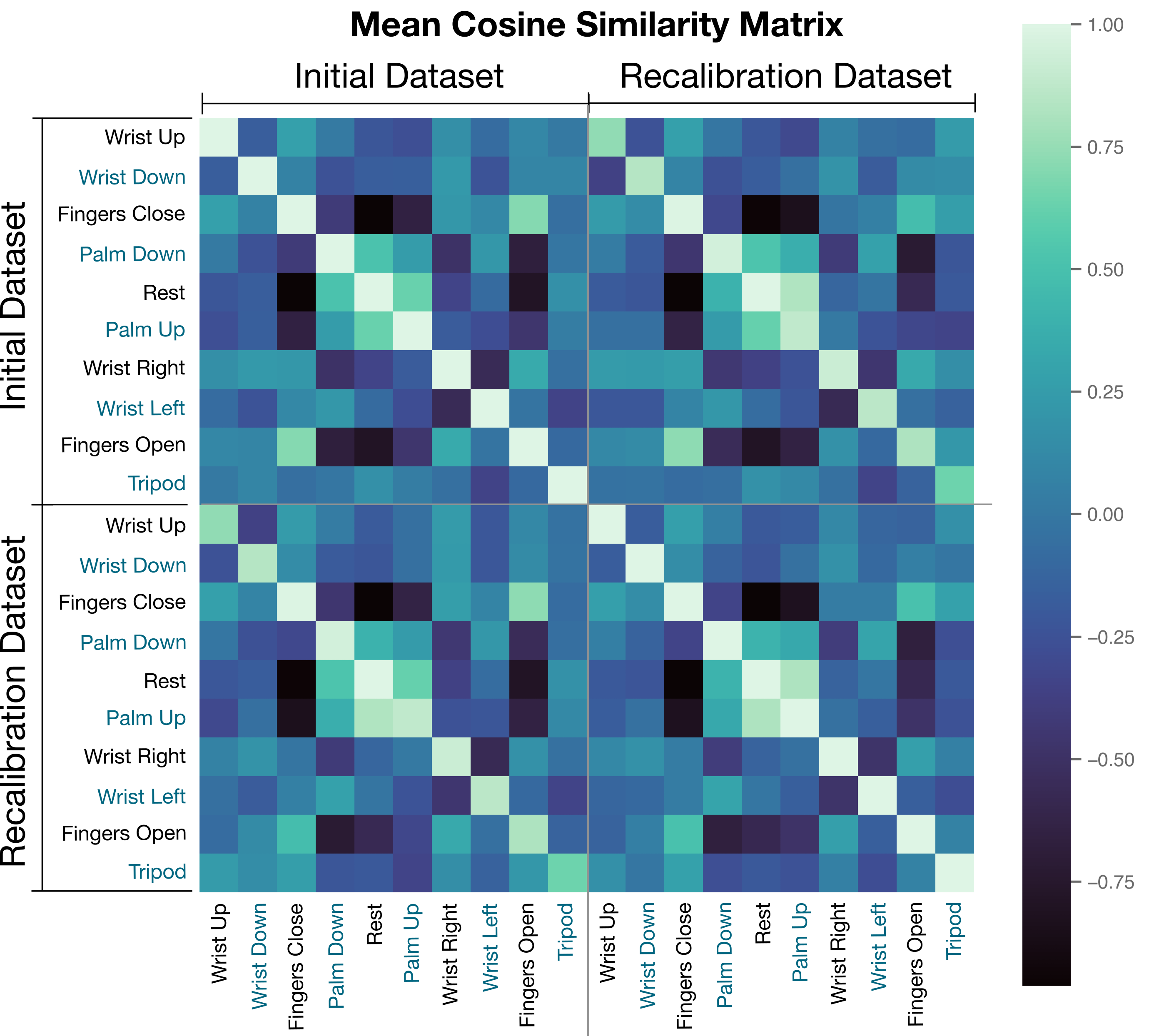}
\caption{\textbf{Comparison of Initial and Recalibration Data. }\textbf{(A)} We plot the RMS heatmaps over the 64 electrodes for the gestures of the initial dataset versus the recalibration dataset here for P08 for the 10 gestures. \textbf{(B)} We show a Euclidean distance matrix and cosine similarity matrix that shows distances and cosine similarities between the mean RMS heatmaps for all the different types of gestures from data collected for all participants. The data are first normalized by electrode.}
\label{fig:datainitialvsrecalibration}
\end{figure}

 \begin{figure}[!]
    \hspace{-1.8em}
    \includegraphics[width=1.0\textwidth]{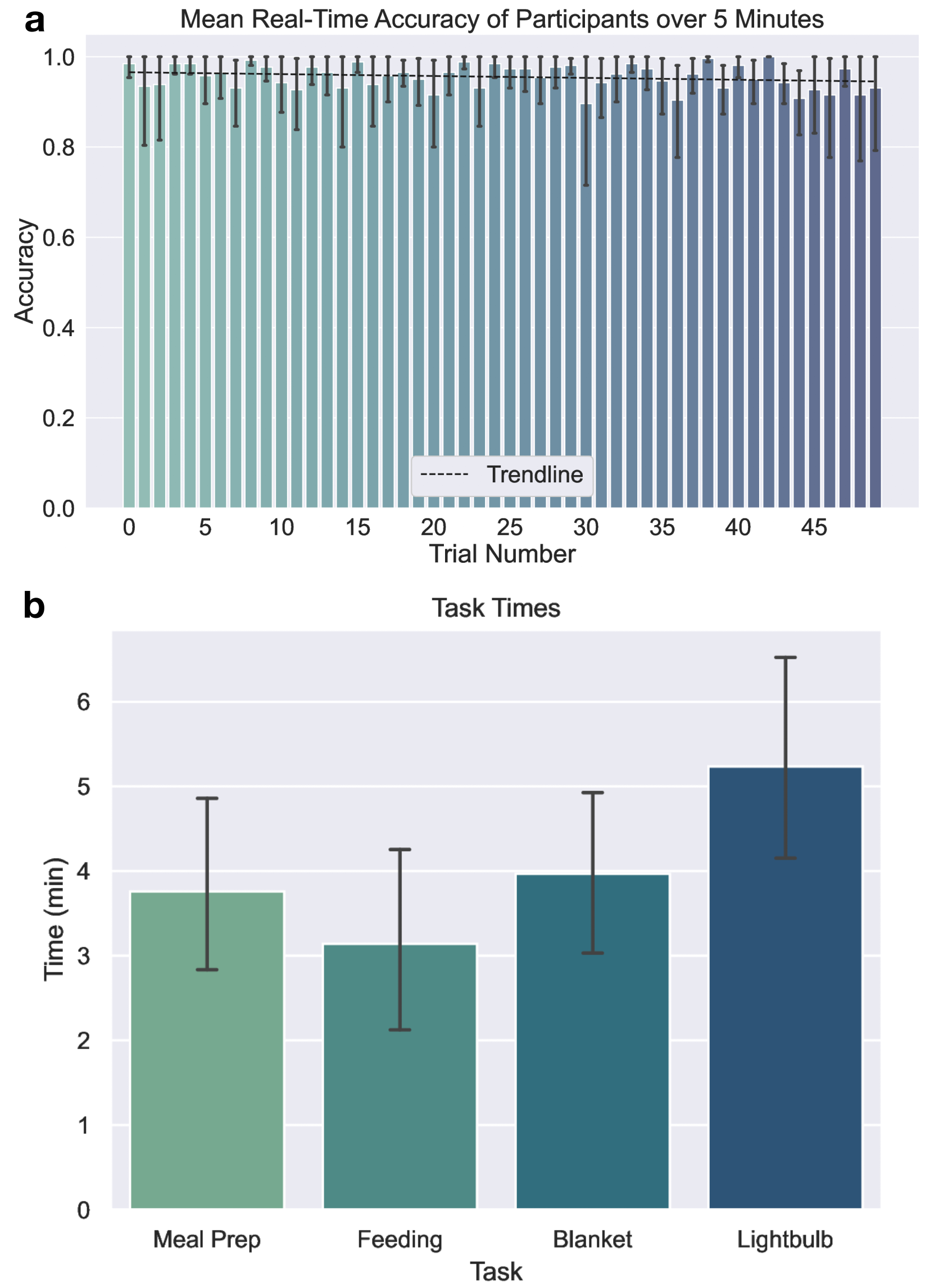}
        
    \caption{\textbf{Real-time Accuracy and Mean Task Times. }\textbf{(A)} Mean real-time classification accuracy for 50 trials where participants are given randomly-ordered cues and visual feedback on the current gesture the classifier is predicting. The trendline of the mean accuracy of participants over time is $y = 0.97 - 4.16 \times 10^{-4}x$. Error bars are given using a 95\% confidence interval of mean participant accuracy. \textbf{(B)} Mean task times for participants in doing various tasks, which shows that mean task times are all below 6 minutes. The error bar shows a 95\% confidence interval for mean task times. } 

    \label{onlineclassification}
\end{figure}

\begin{figure*}[!]
    \includegraphics[width=1.0\textwidth]{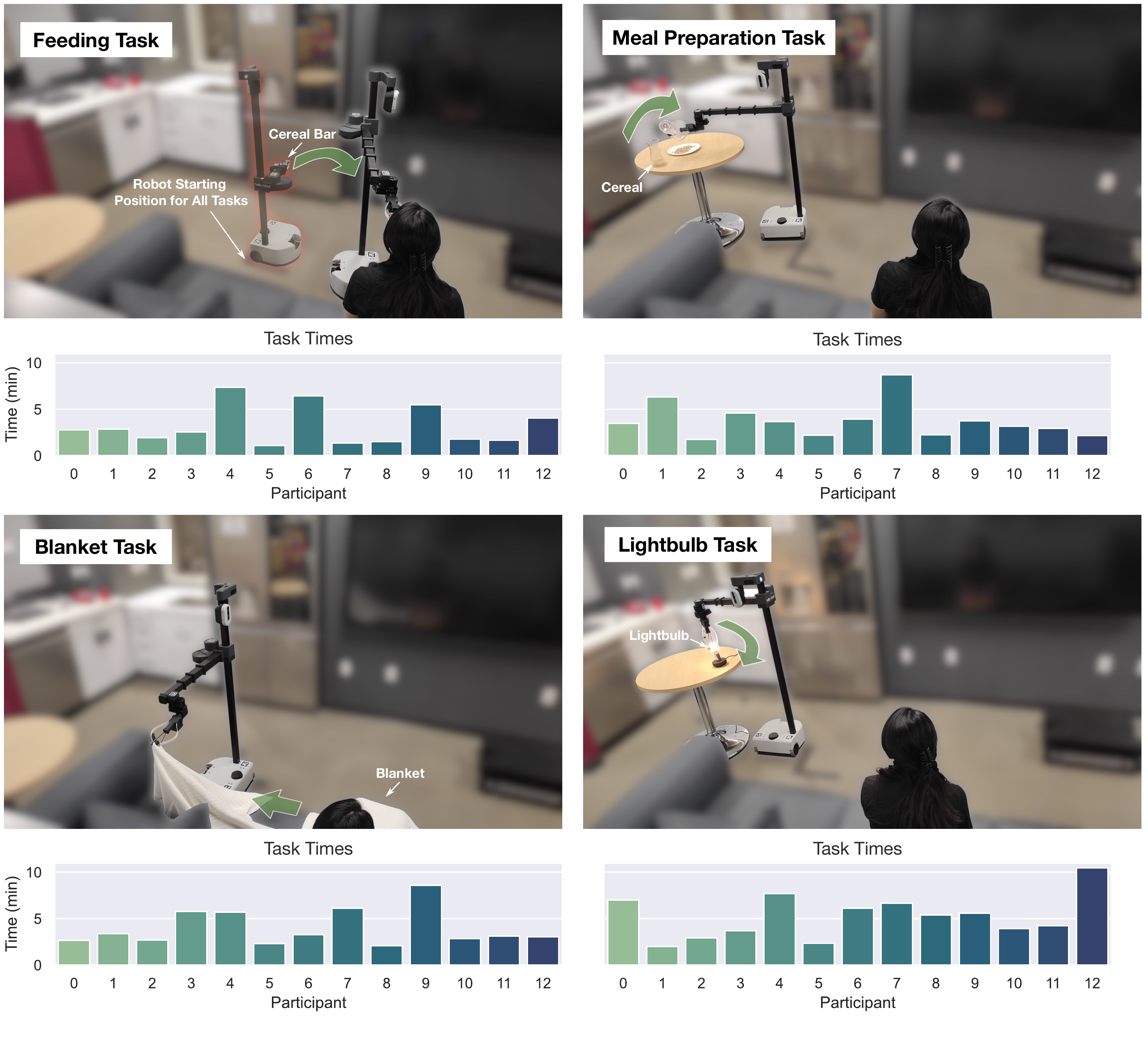}
    \caption{\textbf{Task Setups and Individual Times. }Four different assistive tasks are performed. The first one involves feeding through bringing a cereal bar to the participant's face and the participant taking a bite out of the cereal bar. The second task involves meal preparation, where a carafe of cereal is poured onto a plate. The third task involves blanket manipulation, involving a blanket on the participant to be removed. The fourth task involves screwing in a partially screwed-in lightbulb.} 
    \label{tasksetups}
\end{figure*}

\begin{figure}[!]
    \begin{subcaptiongroup}
        
    \textbf{A}
    
\includegraphics[width=0.6\columnwidth]{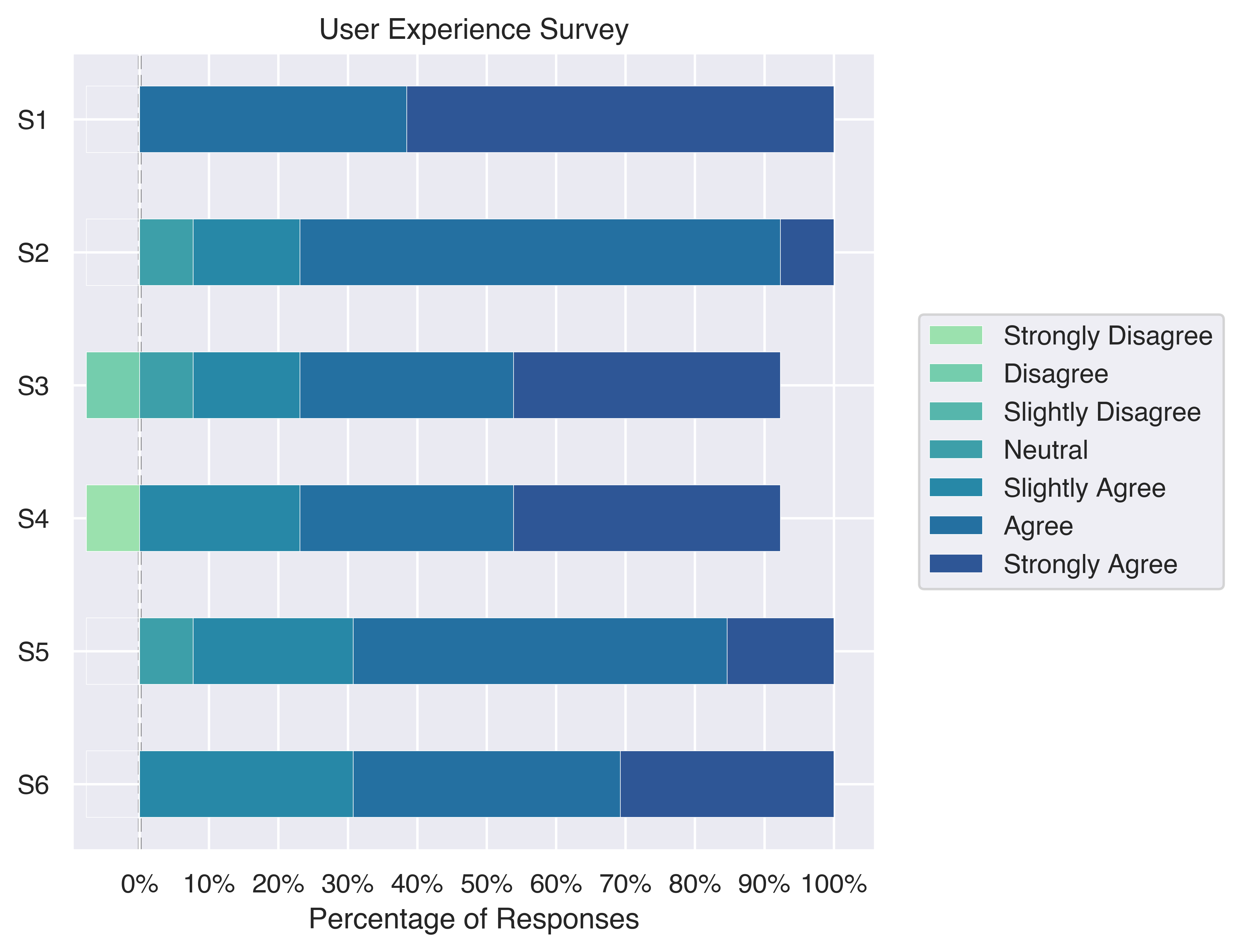} 
    
    \hspace{1ex}\scriptsize{\textsf{S1: I was able to improve my ability to control the robot over time.}}    

    \hspace{1ex}\scriptsize{\textsf{S2: I was able to control the robot effectively using the control interface.}}    

    \hspace{1ex}\scriptsize{\textsf{S3: I find what I am doing with the robot to be intuitive.}}    

    \hspace{1ex}\scriptsize{\textsf{S4: The control interface was easy to learn.}}

    \hspace{1ex}\scriptsize{\textsf{S5: The control interface enabled control of the robot in a reasonable amount of time.}}

    \hspace{1ex}\scriptsize{\textsf{S6: The control interface allowed me to convey my intentions to control the robot.}}
\vspace{1em}

    \textbf{B} 
    
    \includegraphics[width=0.45\columnwidth]{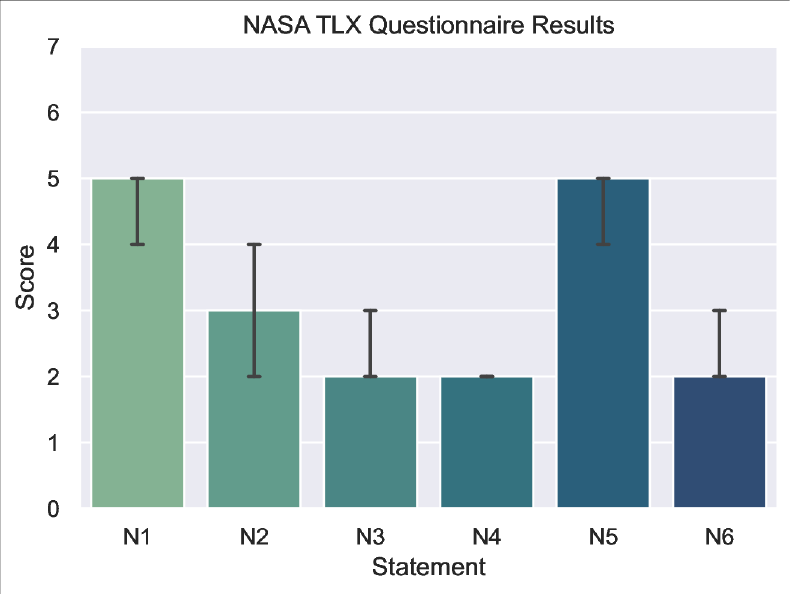} 
    
    \hspace{1ex}\scriptsize{\textsf{N1: How mentally demanding were the tasks?}}    

    \hspace{1ex}\scriptsize{\textsf{N2: How physically demanding were the tasks?}}    

    \hspace{1ex}\scriptsize{\textsf{N3: How hurried or rushed was the pace of the tasks?}}    

    \hspace{1ex}\scriptsize{\textsf{N4: How successful were you in accomplishing what you were asked to do?}}

    \hspace{1ex}\scriptsize{\textsf{N5: How hard did you have to work to accomplish your level of performance?}}

    \hspace{1ex}\scriptsize{\textsf{N6: How insecure, discouraged, irritated, stressed, and annoyed were you?}}
        
    \end{subcaptiongroup}
    \vspace{1em}
    \caption{\textbf{Likert Items and NASA TLX. }\textbf{(A)} Seven-point Likert item responses to statements related to the use of the interface for robot control. The majority of participants agree with the 7 statements about the interface. \textbf{(B)} Seven-point NASA TLX item responses to questions related to the use of the interface for robot control. The closer the score is to 1, the better the score. The max score is 7, which is the worst score. The barplots show the median score, along with an error bar showing a 95\% confidence interval for the median. When controlling the robot to accomplish tasks using our interface, the participants still generally found the interface to be intuitive and easy-to-learn.}
    \label{likertitems}
\end{figure}

\begin{figure}[!]
  \centering
  \includegraphics[width=1.0\columnwidth]{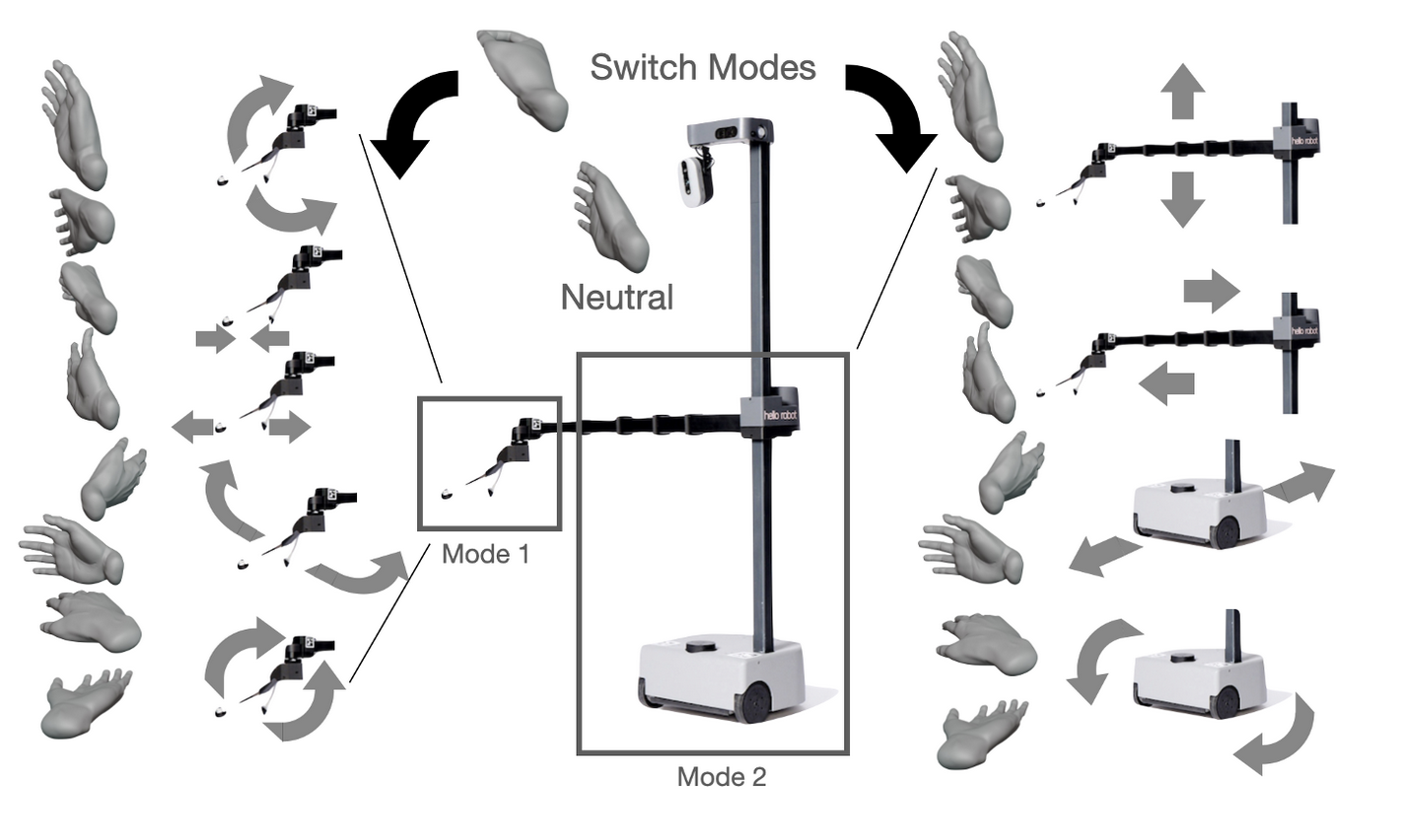}
  \vspace{-10.0pt}
  \caption{\textbf{Robot-Gesture Mapping. }Gestures corresponding to robot degrees of freedom, along with a mode switch gesture command to switch from a) fine wrist and gripper control mode to b) driving and arm control mode. The tripod position can be viewed as a mode switch method, similar to a gear shift. The other gestures are used for directly driving the various motors of the robot.}
  \label{gestures}
\end{figure} 

\clearpage

\appendix

\section*{Supplementary Information}

\renewcommand{\thefigure}{\arabic{figure}}
\captionsetup[figure]{labelsep=colon, name=Supplementary Figure}
\setcounter{figure}{0}

\renewcommand{\thetable}{\arabic{table}}
\captionsetup[table]{labelsep=colon, name=Supplementary Table}
\setcounter{table}{0} % Reset table counter

\setcounter{page}{1}

\begin{figure}[htb!]
  \centering
  \includegraphics[width=1\columnwidth]{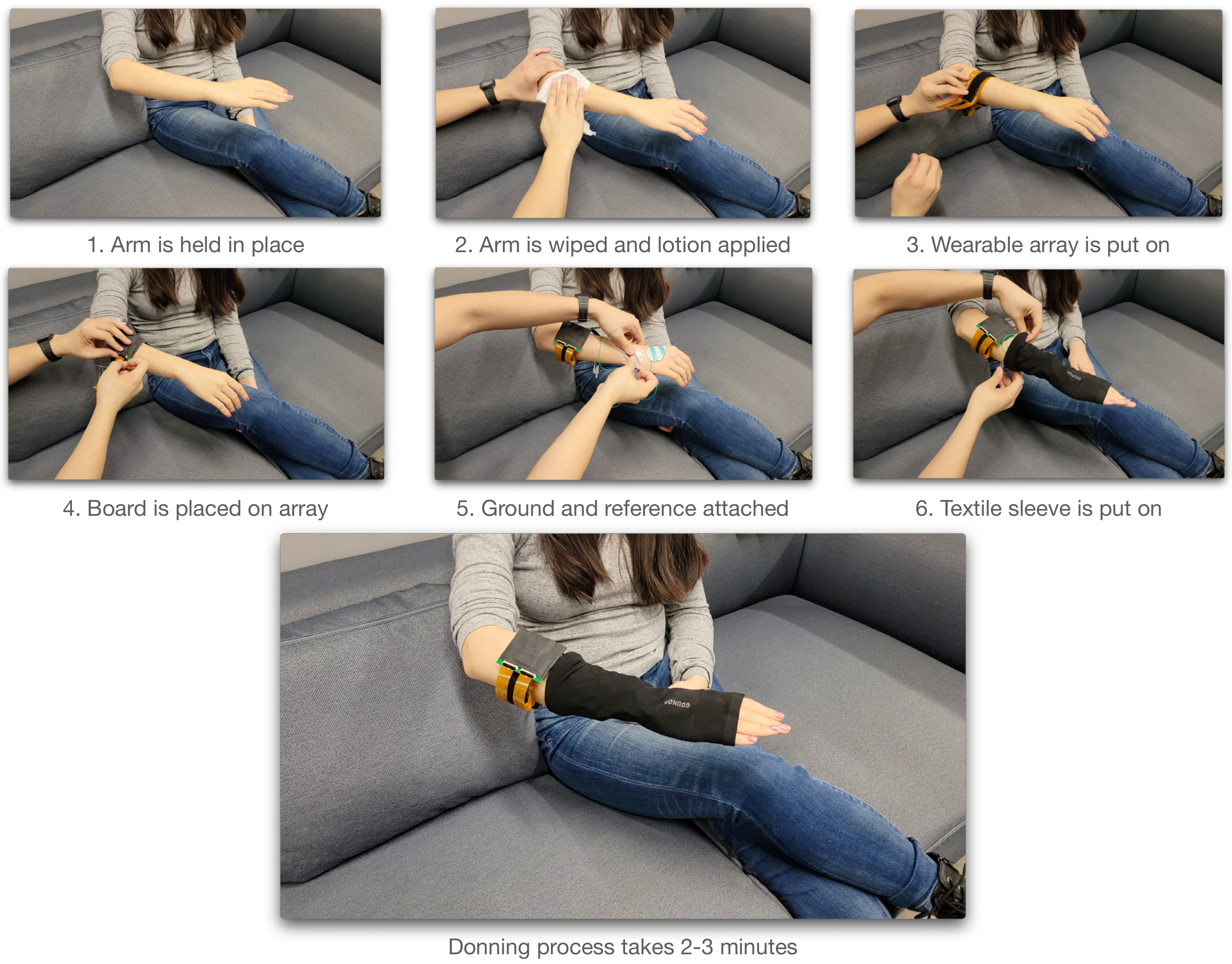}
  \caption{\textbf{Donning Procedure.} The donning process for the wearable is shown here. This procedure takes 2-3 minutes for the full donning process. }
  \label{supfig:donning}
\end{figure}

\begin{figure}[htb!]
  \centering
  \includegraphics[width=0.7\columnwidth]{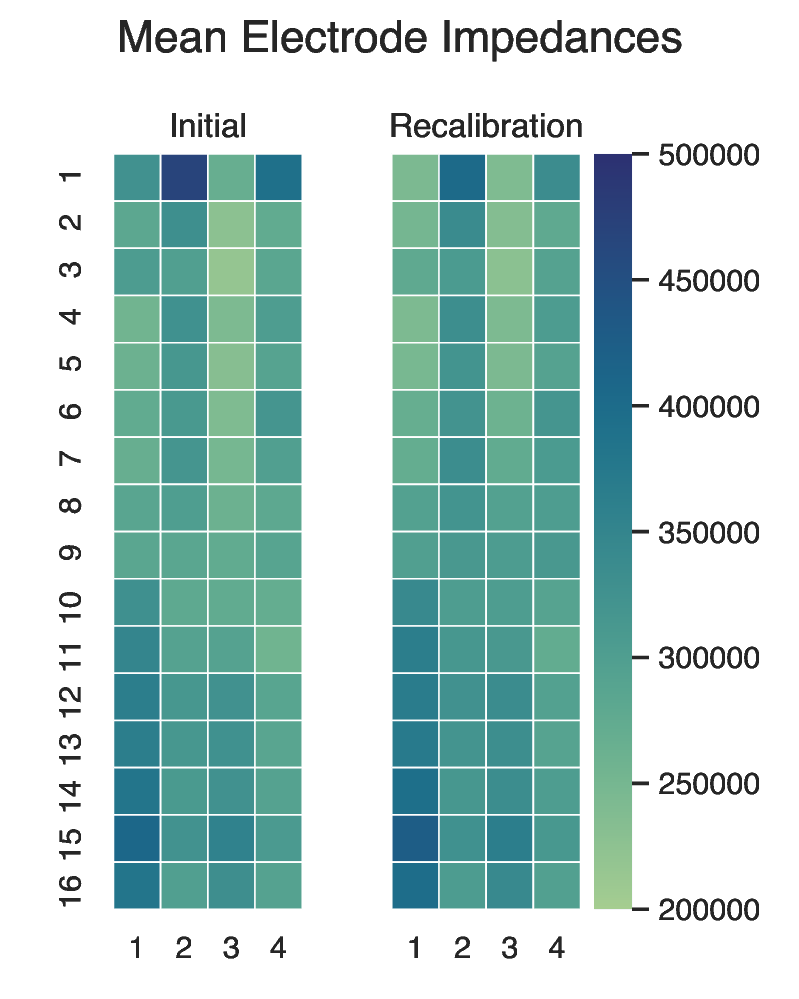}
  \caption{\textbf{Electrode Impedances.} The average impedances of electrodes in Ohms during the initial data collection phase and during the recalibration data collection phase is visualized using these heatmaps.}
  \label{supfig:impedances}
\end{figure}

\begin{figure}[htb!]
  \centering
  \includegraphics[width=1\columnwidth]{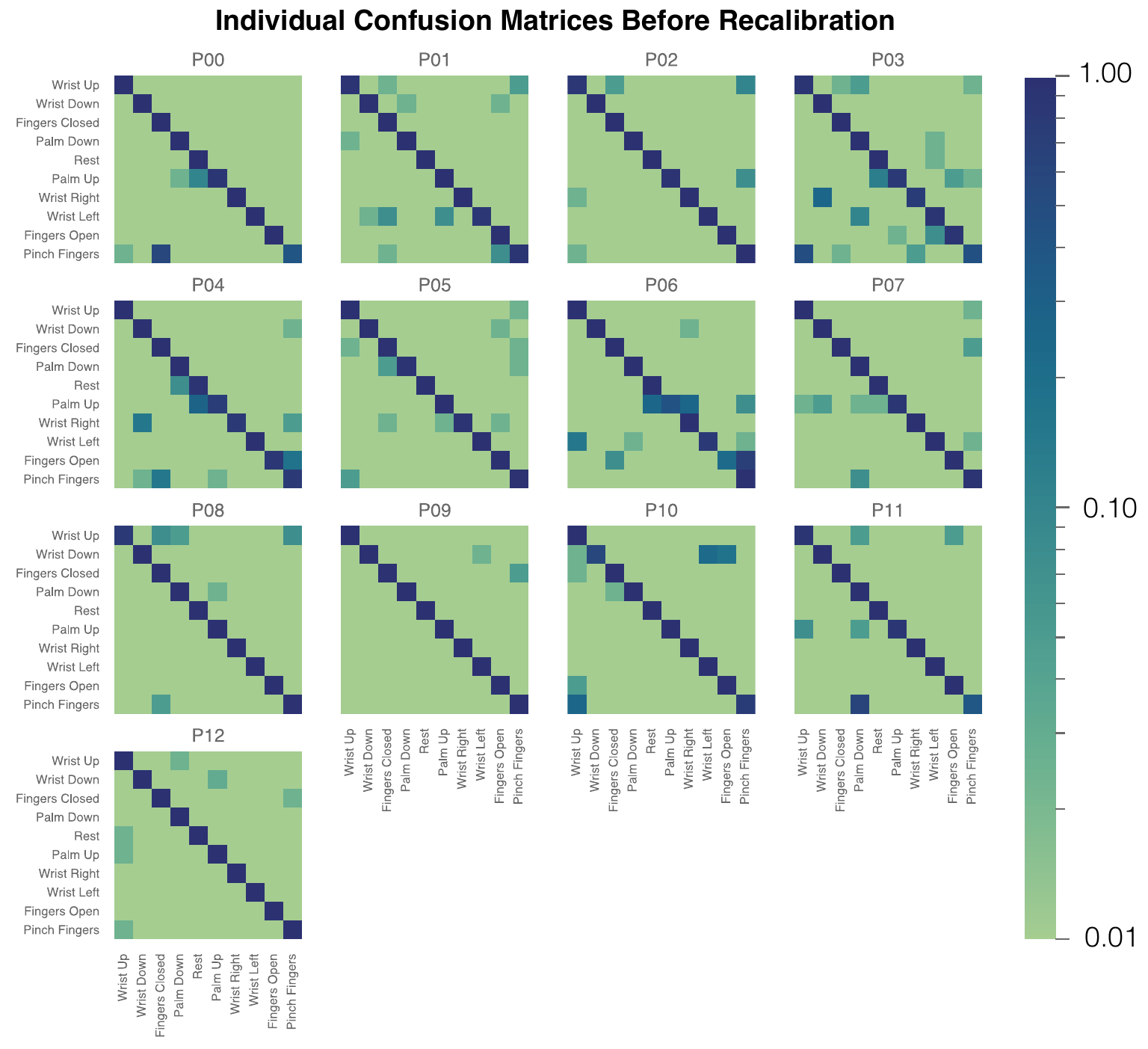}
  \caption{\textbf{Individual Confusion Matrices.} The confusion matrices of individual participants before recalibration is visualized here for participants P00 to P12.}
  \label{supfig:confusion_matrices_before_recalibration}
\end{figure}

\begin{figure}[htb!]
  \centering
  \includegraphics[width=1\columnwidth]{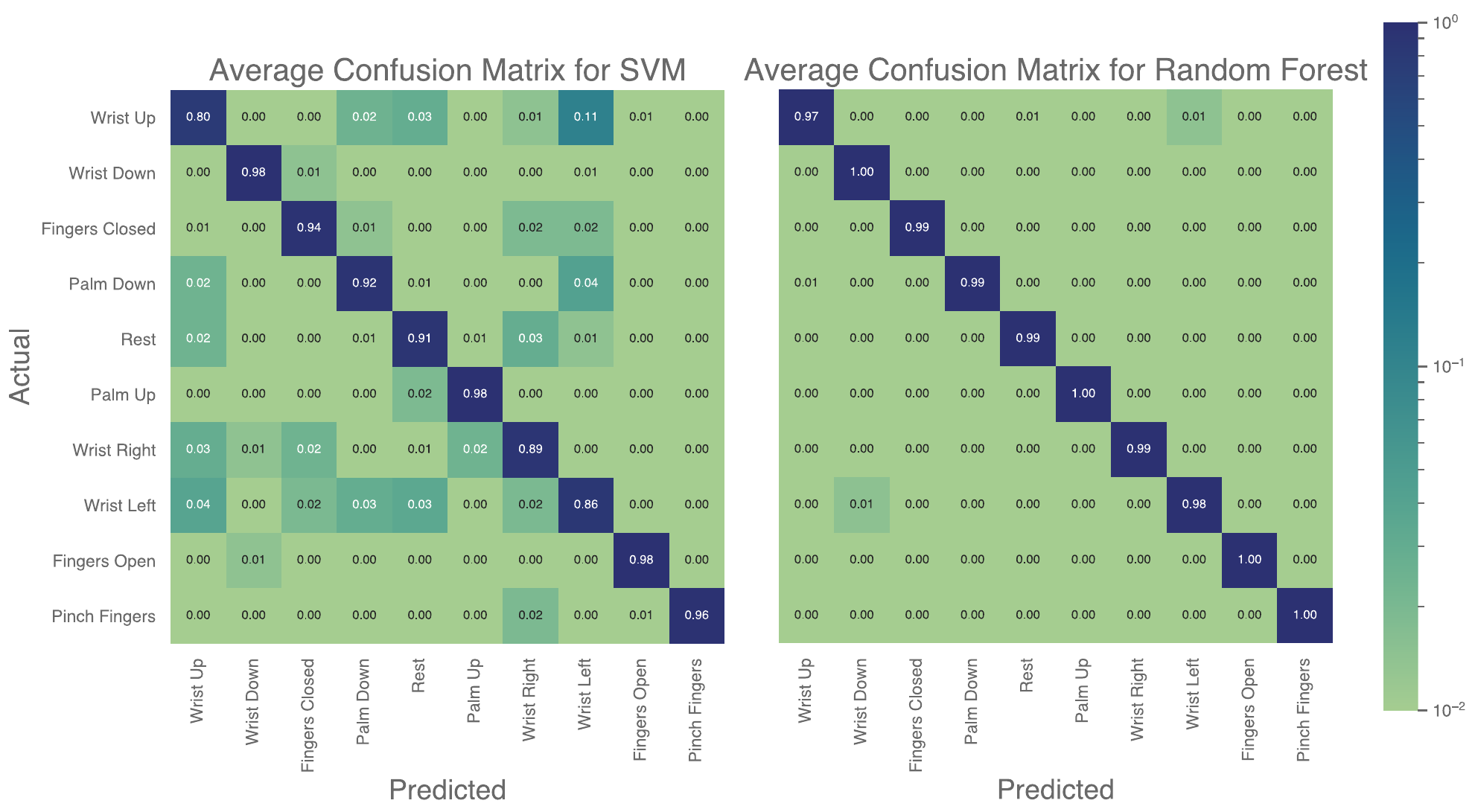}
  \caption{\textbf{Average Confusion Matrices for SVM and RF.} The average test confusion matrices for the SVM and RF models is visualized here. These models are trained and tested on the initial data collection.}
  \label{supfig:svm_rf_confusion_matrices}
\end{figure}

\begin{figure}[htb!]
  \centering
  \includegraphics[width=1\columnwidth]{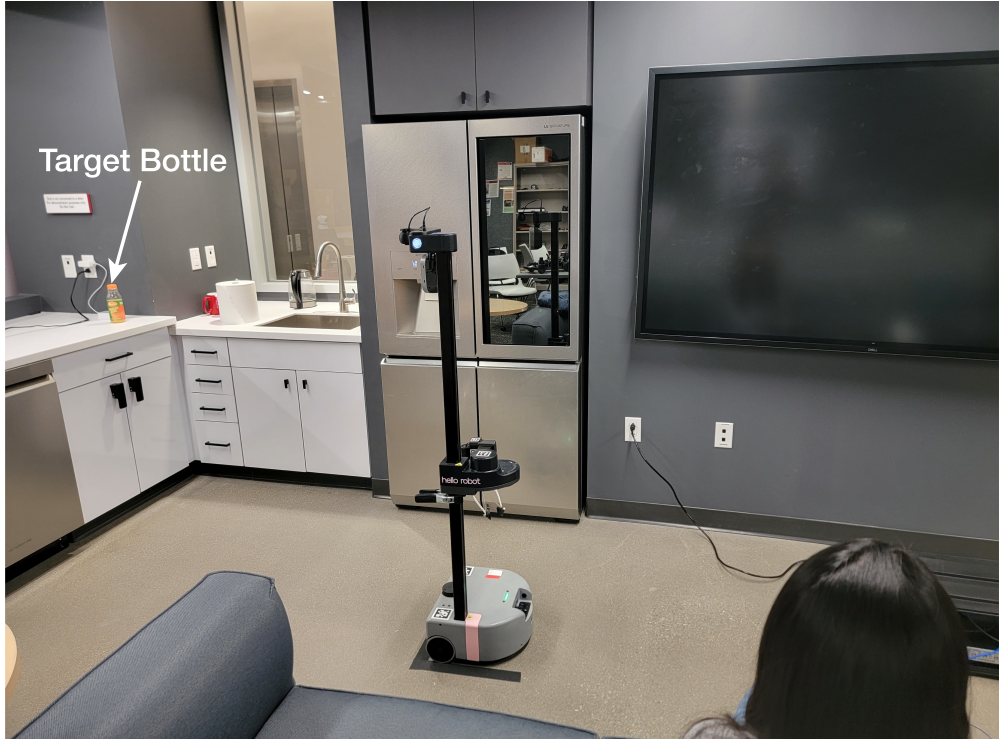}
  \caption{\textbf{Practice Task Setup.} The Setup for the task used for practice, where the bottle of orange juice is grabbed by the mobile manipulator and brought back to the participant is illustrated here.}
  \label{supfig:practicetask}
\end{figure}

\begin{figure}[htb!]
  \centering
  \includegraphics[width=1\columnwidth]{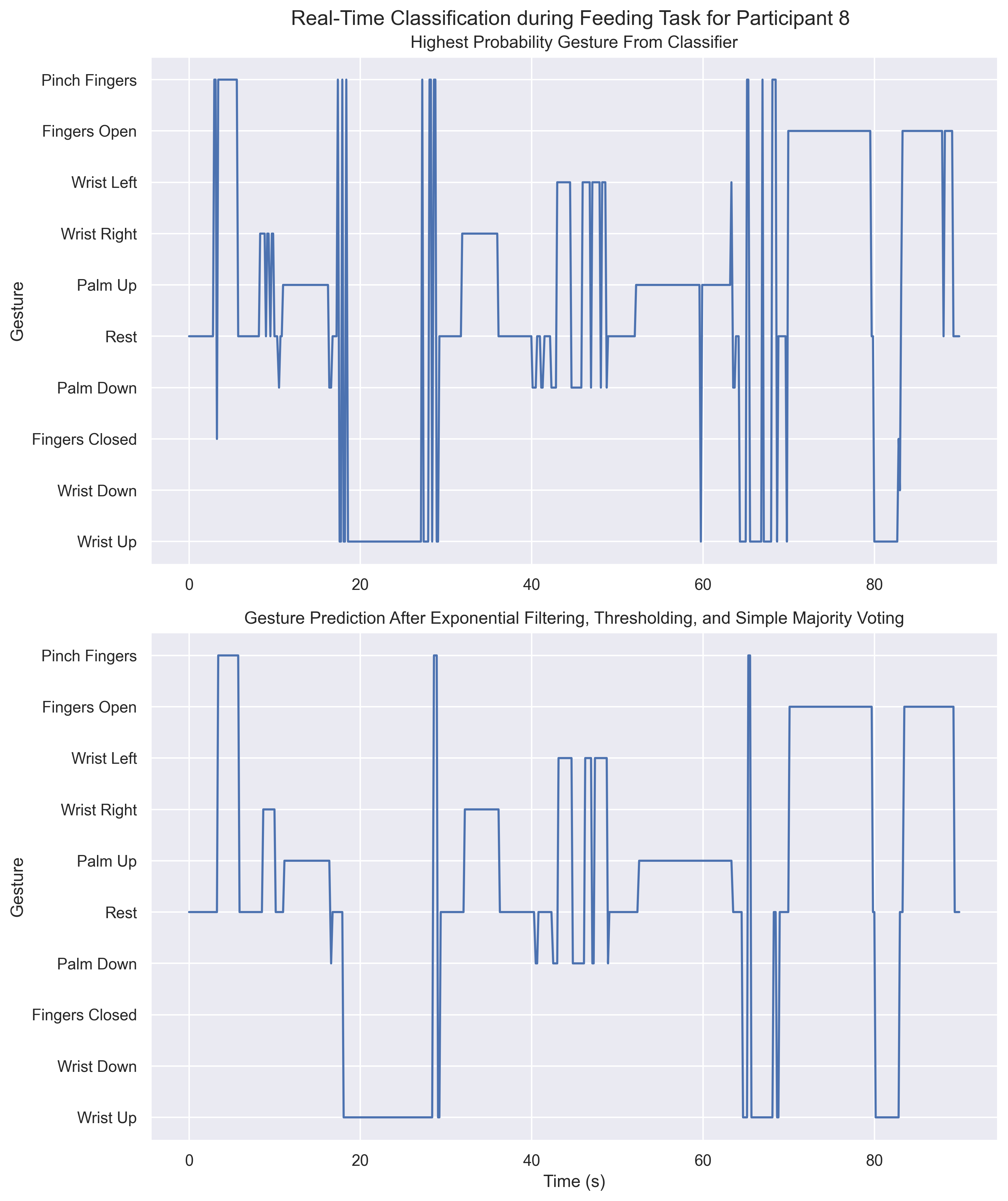}
  \caption{\textbf{Decoder Output.} The entire decoder output for the feeding task for P08 is visualized here. The above plot shows the most likely gesture directly from the decoder, and the below plot shows the gesture prediction after exponential filtering of the decoder output, thresholding, and simple majority voting.}
  \label{supfig:decoder-output}
\end{figure}

\begin{table}[htb!]
\centering
\caption{\textbf{Number of Mode Switches by Participants.}}
\label{suptab:mode_switches}
\sisetup{table-format=2.0}
\begin{tabular}{
  l
  *{7}{S}
}
\toprule & \multicolumn{7}{c}{Participants} \\
\cmidrule(lr){2-8}
{\textbf{Activity}} & {\textbf{P00}} & {\textbf{P01}} & {\textbf{P02}} & {\textbf{P03}} & {\textbf{P04}} & {\textbf{P05}} & {\textbf{P06}} \\
\midrule
Meal Prep  & 9 & 8 & 4 & 12 & 6 & 6 & 13 \\
Feeding  & 1 & 1 & 0 & 3 & 11 & 1 & 23 \\
Blanket Manipulation  & 3 & 3 & 5 & 5 & 7 & 5 & 7 \\
Lightbulb Turning  & 13 & 2 & 8 & 6 & 10 & 4 & 10 \\
\end{tabular}
\sisetup{table-format=2.0}
\begin{tabular}{
  l
  *{6}{S}
}
\toprule
{\textbf{Activity}} & {\textbf{P07}} & {\textbf{P08}} & {\textbf{P09}} & {\textbf{P10}} & {\textbf{P11}} & {\textbf{P12}} \\
\midrule
Meal Prep  & 6 & 2 & 10 & 6 & 6 & 8 \\
Feeding  & 1 & 1 & 7 & 3 & 1 & 3 \\
Blanket Manipulation  & 15 & 3 & 11 & 6 & 4 & 11 \\
Lightbulb Turning  & 10 & 4 & 10 & 10 & 8 & 14 \\
\bottomrule
\end{tabular}
\end{table}

\end{document}